\documentclass[preprint,12pt]{elsarticle}
\usepackage[pdfborder={0 0 0}]{hyperref}
\usepackage{amsthm,amsmath,amsfonts,latexsym,amssymb}
\usepackage{graphics,fancyhdr,graphicx}
\usepackage[ruled,linesnumbered]{algorithm2e}
\usepackage[margin=1in]{geometry}
\usepackage{multirow,booktabs}
\usepackage[title]{appendix}
\usepackage[table]{xcolor}
\usepackage{listings}
\usepackage{tabularx}
\usepackage{placeins}
\usepackage{comment}
\usepackage{lineno}
\usepackage{lscape}
\usepackage{bm}
\usepackage{caption}
\usepackage{subcaption}
\usepackage{float}
\usepackage{soul}
\hypersetup{
    colorlinks = true,
    urlcolor   = blue,
    citecolor  = custom-blue,
}
\usepackage{textgreek}
\usepackage[utf8]{inputenc}
\usepackage[T1]{fontenc}
\biboptions{numbers,sort&compress}
\definecolor{listinggray}{gray}{0.9}
\definecolor{lbcolor}{rgb}{0.9,0.9,0.9}
\definecolor{Darkgreen}{RGB}{0,100,0}

\def \yb{\bm{y}}
\emergencystretch=\maxdimen
\begin{document}
\abovedisplayskip=6.0pt
\belowdisplayskip=6.0pt
\begin{frontmatter}

\title{
Neural operator learning of heterogeneous mechanobiological insults contributing to aortic aneurysms
}

\author[1]{Somdatta Goswami\fnref{fn1}}
\author[2]{David S. Li\fnref{fn1}}
\author[2]{Bruno V. Rego}
\author[3]{Marcos Latorre}
\author[2]{\\Jay D. Humphrey\corref{mycorrespondingauthor}}
\author[1,4]{George Em Karniadakis\corref{mycorrespondingauthor}}

\fntext[fn1]{These authors contributed equally to this work.}
\cortext[mycorrespondingauthor]{Corresponding author. Email: jay.humphrey@yale.edu, george\_karniadakis@brown.edu}

\address[1]{Division of Applied Mathematics, Brown University, Providence, RI, USA}
\address[2]{Department of Biomedical Engineering, Yale University, New Haven, CT, USA}
\address[3]{Center for Research and Innovation in Bioengineering, Valencia Polytechnic University, Valencia, Spain}
\address[4]{School of Engineering, Brown University, Providence, RI, USA}

\begin{abstract}
\noindent Thoracic aortic aneurysm (TAA) is a localized dilatation of the aorta resulting from compromised wall composition, structure, and function, which can lead to life-threatening dissection or rupture. Several genetic mutations and predisposing factors that contribute to TAA have been studied in mouse models to characterize specific changes in aortic microstructure and material properties that result from a wide range of mechanobiological insults. By contrast, assessments of TAA progression \emph{in vivo} are largely limited to measurements of aneurysm size and growth rate. It has been shown, however, that aortic geometry alone is not sufficient to predict the patient-specific progression of TAA but computational modeling of the evolving biomechanics of the aorta could predict future geometry and properties from initiating insults. 
In this work, we present an integrated framework to train a deep operator network (DeepONet)-based surrogate model to identify contributing factors for TAA by using synthetic finite element-based datasets of aortic growth and remodeling (G\&R) resulting from prescribed mechanobiological insults. For training data, we investigate multiple types of TAA risk factors and spatial distributions within a computationally efficient constrained mixture model to generate axial--azimuthal maps of aortic dilatation and distensibility. The trained network is then capable of predicting the initial distribution and extent of the insult from a given set of dilatation and distensibility information, which in turn can be used to determine subsequent aortic geometry and mechanical properties. Two DeepONet frameworks are proposed, one trained on sparse information and one on full-field grayscale images, to gain insight into a preferred neural operator-based approach. Performance of the surrogate models is evaluated through multiple simulations carried out on insult distributions varying from fusiform (analytically defined) to complex (randomly generated). We show that this integrated continuous learning modeling approach can predict the patient-specific mechanobiological insult profile associated with any given dilatation and distensibility map with a high accuracy, particularly when based on full-field images. Our findings demonstrate the feasibility of applying DeepONet to support transfer learning of patient-specific inputs (e.g., age, hypertension, diabetes, Marfan syndrome) to predict TAA progression with clinical images of the aorta.
\end{abstract}


\end{frontmatter}

\section{Introduction}
\label{sec:intro}

\noindent Thoracic aortic aneurysms (TAAs) are localized dilatations of the aorta that associate with a higher risk of life-threatening aortic dissection or rupture; they can initiate from a variety of biomechanical and genetic factors, often developing over several years \cite{Lindsay2014, Ramirez2018, FaggionVinholo2019, Milewicz2019, Pinard2019, LindquistLiljeqvist2021}. Treatment of TAAs may involve surgical replacement with a synthetic graft or repair via the placement of an endovascular stent \cite{Zoli2010, Ziza2016}. Determination of the optimal approach and time of intervention depends on myriad factors, including the size, growth rate and location of the aneurysm, as well as the individual's genetic history \cite{Pirruccello2022}; ultimately, to facilitate systematic improvement of the patient's prognosis and therapeutic design, there is a pressing need to understand the complex roles these aspects play in the development of aneurysms \cite{Milewicz2017}. 

Assessment of aortic health in the clinic is largely limited to \emph{in vivo} anatomical information and hemodynamic measurements, with few biomarkers available. On the other hand, murine models of aortic aneurysm have provided valuable insight into the structural and biomechanical properties of the normal and diseased thoracic aorta via \emph{in vitro} experimentation of excised tissue specimens \cite{Bellini2016, Cavinato2021} that complement \emph{in vivo} studies. Several genetic mutations leading to compromised aortic structure and function have been identified as critical predisposing factors driving TAA formation, including those affecting extracellular matrix integrity, smooth muscle contractile dysfunction, and aberrant intracellular signaling \cite{Lima2010, Humphrey2015, Ramirez2018, estrada2021roles}. Although much has been learned from these studies, multiple contributors are often present in combination \emph{in vivo}, rendering it challenging to identify relationships among different mechanisms. Nevertheless, it is clear that characterization of the biomechanics of the aorta is necessary to gain a deeper knowledge of aortic disease progression. Toward this end, computational models of aneurysm growth and remodeling can facilitate mechanistic understanding by isolating the influence of individual biomechanical defects on subsequent aneurysm progression \cite{Valentin2009b, Wilson2013, Liang2017, weiss2021biomechanical}. A theoretical model with an ability to generate robust predictions of aneurysm progression from limited anatomical information can both generate synthetic datasets for analysis and lay the foundation for enriched diagnosis of aortic function beyond clinically available measurements.

Current advancements in modeling now provide the opportunity to leverage machine learning, which has emerged as an effective surrogate model for high-fidelity solvers, in order to overcome previous computational hurdles that would otherwise make such modeling intractable for clinically relevant time frames. Such surrogate models have demonstrated the potential for automated measurement of aortic geometry, patient risk stratification, and the prediction of aneurysm growth and rupture \cite{Pradella2021, Ostberg2022, he2021estimating, Liu2021, Zhou2021, LindquistLiljeqvist2021}. Additionally, physics-informed neural networks (PINNs) \cite{raissi2019physics, goswami2020adaptive_dem, goswami2020transfer, jagtap2020conservative} have been a promising advancement in the domain of scientific machine learning. However, to simulate multiple initial/boundary problems having different applied loading, one often needs to retrain a PINN. Hence, developing models that can learn the operator-level mapping between functions (that is, forecasting the physical system under a variety of initial/boundary circumstances) is critical \cite{goswami2022physics, yin2022simulating}. In this case, neural operators can learn nonlinear mappings between function spaces, providing a novel simulation paradigm for real-time prediction of complicated dynamics. A deep operator-based neural network (DeepONet) proposed in \cite{lu2021learning} is now popular for learning solutions from labelled input-output datasets consisting of varied initial/boundary conditions and different forcing functions. The idea of DeepONet is motivated by the Universal Approximation Theorem for Operators, which defines a new and relatively underexplored realm for deep neural network (DNN)-based approaches that map infinite-dimensional functional spaces rather than finite-dimensional vector spaces (functional regression). The computational model consists of two classes of DNNs; one encodes the input function at fixed sensor points (branch net) while the other accounts for the location of the output function (trunk net). 

Here, we present an integrated computational platform for predicting contributing factors to TAA by melding a constrained mixture model of TAA enlargement with DeepONet. Specifically, we use a previously established computationally efficient implementation of our mechanobiologically equilibrated constrained mixture model to describe the long-term evolution of TAAs. We propose two surrogate models to approximate aneurysmal initiating factors: the first framework admits sparse information to be encoded in the branch network, while the second instead encodes full-field grayscale images taking into account patient-specific characteristics \cite{wojnarski2018machine}. The accuracy of each approach is evaluated for multiple types of simulated aneurysms arising from multiple contributors investigated in past studies, and several variations of the neural network design are explored to compare the relative performance of different architectures. Our findings demonstrate feasibility of this technique for future analysis of clinical images of the aorta. The key highlights are:

\begin{itemize}
    \item A novel framework to predict TAA pathology melds a constrained mixture model for arterial growth and remodeling with a DeepONet-based surrogate model.
    \item 3D finite element simulations of TAA progression arise from randomly distributed losses of elastic fiber integrity and dysfunctional mechanosensing.
    \item The generalizable DeepONet can predict the solution with sufficient accuracy, even when provided limited information.
    \item Performance is improved by employing convolutional neural networks rather than fully connected feed-forward neural networks with sparse information.
    \item The preferred network architecture takes as input grayscale images of dilatation and distensibility to predict the insult profile.
    \item This approach can be extended to patient-specific medical images to provide patient-specific solutions.
\end{itemize}

\section{Methods}
\label{sec:methods}

\begin{figure}[!htbp]
\centering
\includegraphics[width=0.85\textwidth]{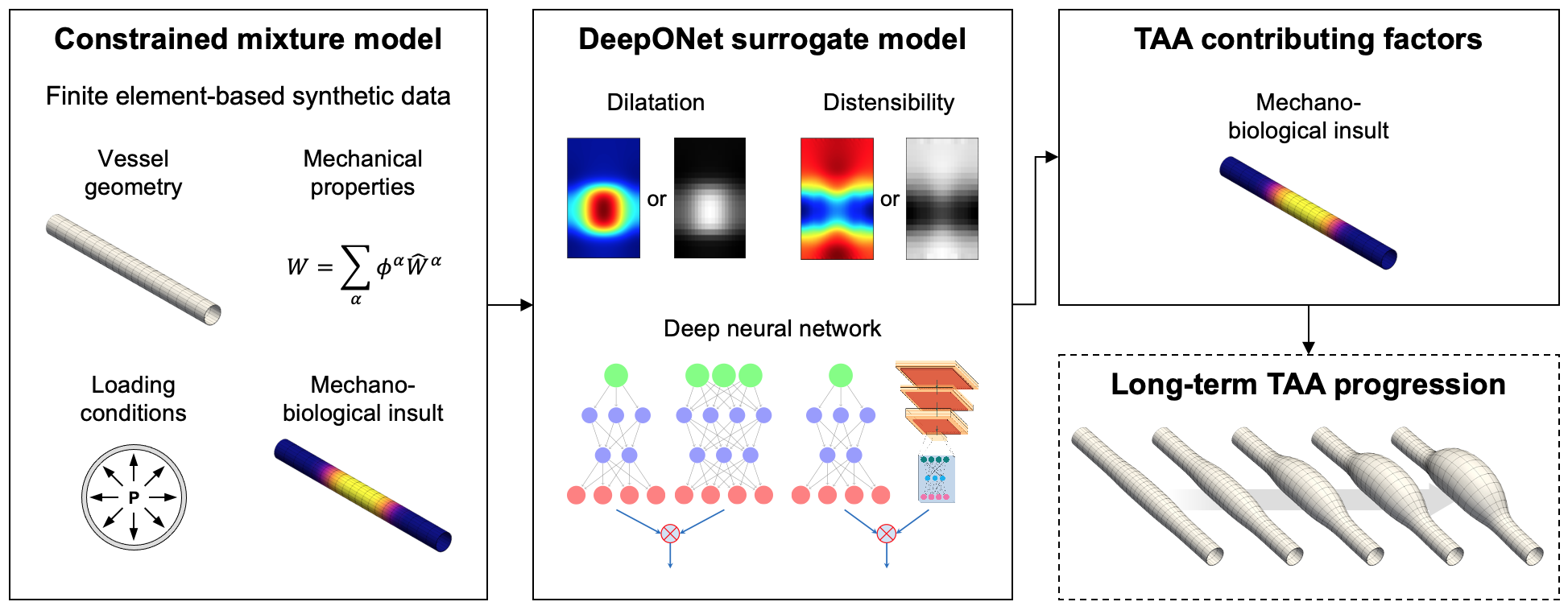}
\caption{Pipeline for training the DeepONet to predict mechanobiological insults for forecasting long-term aneurysm growth. First, 3D finite element simulations of TAA evolution incorporating vessel geometry, mechanical properties, \emph{in vivo} loading conditions, and mechanobiological insult profiles are used to generate synthetic training and testing data for the DeepONet surrogate model, which is comprised of a deep neural network with associated trainable parameters and evaluation points. Subsequently, the DeepONet is trained using labeled maps of local dilatation and distensibility corresponding to the prescribed insults, as either axial--azimuthal maps or converted grayscale images. Finally, the insult profile (distribution and severity) is predicted from a given set of dilatation and distensibility information, which can be related to estimation of TAA progression.}
\label{fig:flowchart}
\end{figure}

\noindent To generate synthetic data for training the surrogate model, we employ a previously established constrained mixture model within a finite element platform \cite{Latorre2020b}. Finite element simulations incorporating the initial geometry of the aorta, bulk mechanical properties of the vessel wall, and \emph{in vivo} loading conditions (axial pre-stretch and intraluminal blood pressure), along with a prescribed mechanobiological insult to initiate TAA, allow predictions of local dilatation and distensibility fields (\autoref{subsec:taamodel}). These are converted into axial--azimuthal maps and randomly categorized into testing and training sets, which enable training of the DeepONet (\autoref{subsec:deeponet}) to predict the initial insult profile from a given dataset. Finally, prediction of the spatial distribution and severity of the insult profile is used for the projection of future TAA growth. The modeling pipeline is summarized in \autoref{fig:flowchart}.

\subsection{Constrained mixture model for arterial growth and remodeling}
\label{subsec:cmm}

\noindent Over the past several years, we have developed a constrained mixture formulation to describe growth and remodeling (G\&R) of the aortic wall, which has been used to model the changing composition, structure, and mechanics in both animal and clinical studies \cite{Humphrey2002, Baek2006, Valentin2009, Latorre2018}. We have recently implemented a computationally efficient 3D finite element framework for modeling the initiation and long-term growth of TAA within a thick-walled cylindrical vessel segment \cite{Latorre2020b}. In the following, we summarize the pertinent details of the modeling framework; further details can be found in the original papers \cite{Latorre2018, Latorre2020a}.

\subsubsection{Modeling framework}

\noindent We model the arterial wall as a mixture of three primary load-bearing constituents: elastin-dominated matrix $e$, collagen fiber-dominated matrix $c$, and smooth muscle cells $m$, that can undergo changes in mass (grow) and microstructure (remodel) in response to biomechanical stimuli. These constituents $\alpha = e,c,m$ are governed by mass density production rates $m^\alpha(\tau)$ at the time of deposition $\tau$, removal (decay) functions $q^\alpha(s,\tau)$, with $s$ denoting the current time, and stored energy density functions $\hat{W}^\alpha(s,\tau)$ that depend on the constituent-specific deformations relative to their evolving natural configurations $n(\tau)$ and thus describe the mechanical behaviors.

Because the extracellular environment is mechanoregulated by cells within the vessel wall, production and removal rates can be modulated by perturbations from the homeostatic state (i.e., a quasi-equilibrium where production balances removal, denoted by the subscript $o$). Production per unit volume at time $\tau$ is expressed using a potentially evolving nominal rate $m_N^\alpha(\tau) > 0$ that is modulated by deviations in intramural stress $\sigma$ (induced by pressure and axial force) and wall shear stress $\tau_w$ (induced by blood flow) from homeostatic values, namely,

\begin{equation} \label{eq:production}
m^\alpha(\tau) = m_N^\alpha(\tau) \bigg( 1 + K_\sigma^\alpha \Delta\sigma(\tau) - K_{\tau_w}^\alpha \Delta\tau_w(\tau) \bigg) = m_N^\alpha(\tau) \Upsilon^\alpha(\tau),
\end{equation} \smallskip

\noindent where $K_i^\alpha$ are gain-type parameters controlling the sensitivity to stress deviations $\Delta\sigma$ and $\Delta\tau_w$ from set points $\sigma_o$ and $\tau_{wo}$, summarized in the stimulus function $\Upsilon^\alpha$. \autoref{eq:production} is formulated to replicate experimental observations that increased wall stress and increased flow tend to heighten and reduce extracellular matrix production, respectively \cite{Valentin2009b, Latorre2018}. In this study, we define $\Delta\sigma = (\sigma - \sigma_o)/\sigma_o$ and $\Delta\tau_w = (\tau_w - \tau_{wo})/\tau_{wo}$, where $\sigma$ is one-third the trace of the in-plane wall stress. Additionally, we assume that the intraluminal pressure within the aorta remains constant (no pulsatility) when computing the G\&R and that the flow through the aorta remains constant and laminar.

Constituent removal between times $\tau$ and $s$ is modeled by

\begin{equation} \label{eq:removal}
q^\alpha(s,\tau) = \exp \bigg( -\int_\tau^s k_N^\alpha \bigg( 1 + \omega \Big( \Delta\sigma(t) \Big)^2 \bigg) dt \bigg),
\end{equation} \smallskip

\noindent where $k_N^\alpha$ is the basal removal rate, and $\omega > 0$ is a gain-type parameter for intramural stress deviations. Note that, while increases in intramural stress can drive increased production, as stated above, they can also stimulate the removal of constituents through activation of matrix-degrading enzymes, as reflected in \autoref{eq:removal}. With the above definitions, we then express the mass density per unit reference volume of each constituent $\rho_R^\alpha$ by

\begin{equation} \label{eq:massdens}
\rho_R^\alpha(s) = \int_{-\infty}^s m_R^\alpha(\tau) \, q^\alpha(\tau) \, d\tau,
\end{equation} \smallskip

\noindent where $m_R^\alpha = Jm^\alpha$ is the referential (subscript $R$) mass density production rate, with $J = \sum \rho_R^\alpha/\rho$ representing the volume ratio between the reference and current \emph{in vivo} configurations of the tissue constituents.

We employ a hyperelasticity framework to describe the transient biomechanical properties of the vessel wall, which is modeled as a mixture of nonlinear, nearly incompressible, anisotropic materials. This allows the intramural stress-driven changes in production and removal to be defined in terms of stored energy density relations $\hat{W}^\alpha(s,\tau)$, which are functions of the multiaxial deformations of each constituent. The Cauchy stress at the tissue (mixture) level $\bm{\sigma}$ includes passive contributions from each constituent, given by

\begin{equation} \label{eq:Cauchystress}
\bm{\sigma}(s) = -p(s)\mathbf{I} + \frac{2}{J(s)} \, \mathbf{F}(s) \, \frac{\partial W_R (s)}{\partial \mathbf{C}(s)} \, \mathbf{F}^\mathrm{T}(s),
\end{equation} \smallskip

\noindent where $p$ is a Lagrange multiplier that enforces the transient isochoric motions, $\mathbf{I}$ is the identity tensor, $\mathbf{F}$ is the deformation gradient tensor for the tissue from reference to current configurations, $\mathbf{C} = \mathbf{F}^\mathrm{T} \mathbf{F}$ is the right Cauchy-Green tensor, $J = \det \mathbf{F}$, and $W_R = \sum W_R^\alpha$ is the total stored energy of the mixture. The constitutive relations are defined separately for $e$, $c$, and $m$ and are weighted by their respective volume fractions $\phi^\alpha$ ($W_R^\alpha = \phi^\alpha \hat{W}^\alpha$). Further details are described in \autoref{app:appendix_A_constprop}. Note that, because many TAAs associate with diminished or absent smooth muscle contractility, we assume that the stress depends only on passive properties.

\subsubsection{Mechanobiologically equilibrated constrained mixture model}
\label{subsubsec:cmm}

\noindent In cases of TAA, where the characteristic timescale of G\&R is frequently shorter than that of the biomechanical stimulus (e.g., elastic fiber degradation with gradual hypertension), G\&R can be assumed to reach a quasi-static mechanobiological equilibrium at time $s \gg 0$, thus allowing a time-independent approach at which production balances removal $\forall s$ and computation of the heredity integrals is not required. In particular, the stimulus function in \autoref{eq:production} reduces to

\begin{equation}
\Upsilon_{h}^{\alpha}(\Delta \sigma_{h}, \Delta\tau_{wh}) = 1,
\end{equation} \smallskip



\noindent with $h$ denoting the evolved homeostatic state. Accordingly, rule-of-mixtures expressions can be used to be used for stored energy ($W_{Rh} = \sum \phi_{Rh}^\alpha \hat{W}^\alpha$, where $\phi_{Rh}^\alpha$ are the evolved constituent mass fractions). Similarly, for the Cauchy stresses (\autoref{eq:Cauchystress}),

\begin{equation}
\bm{\sigma}_h = -p_h\mathbf{I} + \sum_{\alpha}^{e,c,m} \phi_h^\alpha \hat{\bm{\sigma}}_h^\alpha,
\end{equation} \smallskip

\noindent where $\hat{\bm{\sigma}}_h^\alpha$ are the constituent-specific Cauchy stresses, and $p_h$ is the equilibrated Lagrange multiplier for the quasi-static G\&R evolution (see \autoref{app:appendix_A_mbe} for more details).

\subsection{Modeling TAA growth from prescribed mechanobiological insults}
\label{subsec:taamodel}

\noindent Although TAAs typically exhibit an irregular diameter, eccentricity, and thickness, it remains instructive to consider smoothly varying insult profiles prescribed on an initially straight cylindrical vessel, with \emph{in vivo} geometry and mechanical properties derived from our previous studies (uniform wall thickness $h_o = 40$ \textmu m and luminal radius $r_o = 647$ \textmu m; see \autoref{app:appendix_A_mbe}). Here, aneurysms are initiated by prescribing an insult that emulates defects such as a localized breakage in elastic fibers or disruption at cellular integrin binding sites. We adopt a cylindrical coordinate system in the initial homeostatic state $\{ r_o, \theta_o, z_o \}$, and two approaches for insult profile definitions are considered: analytical and randomly generated.

\subsubsection{Analytically defined insult profiles}

\begin{figure}[!htbp]
\centering
\includegraphics[width=0.75\textwidth]{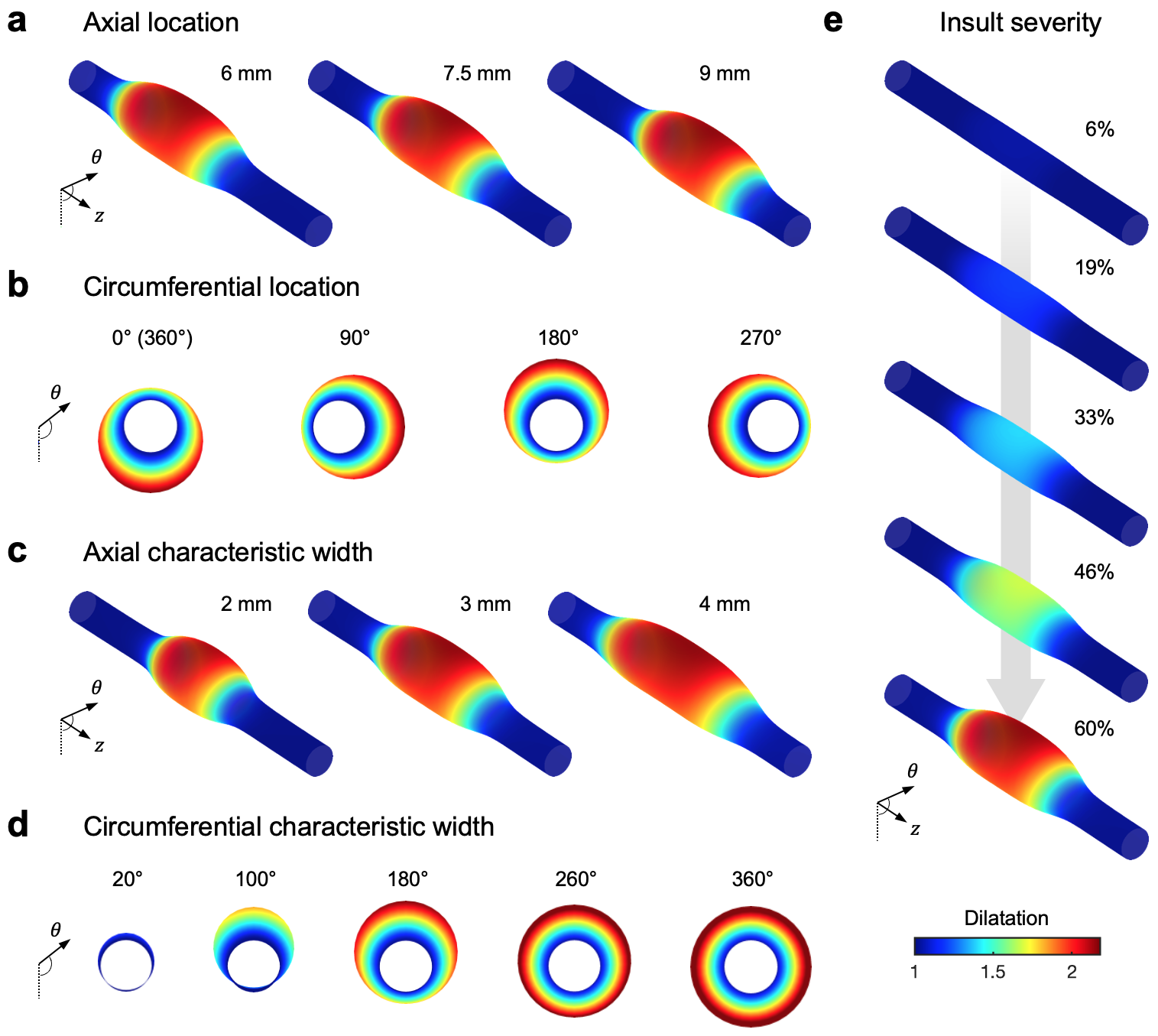}
\caption{Examples of analytically defined insults and their influence on TAA geometry. Dilatation (remodeled inner radius normalized by initial inner radius) maps are shown for variations in the (a) axial location $z_{apex} \in [6,9]$ mm and (b) circumferential location $\theta_{apex} \in [0^\circ, 360^\circ]$ of the aneurysm apex, as well as the aneurysm (c) axial characteristic width $z_{od} \in [2,4]$ mm and (d) circumferential characteristic width $\theta_{od} \in [20^\circ, 360^\circ]$. (e) For each combination of apex location and characteristic width, insults (e.g., loss of elastic fiber integrity) of increasing severity are modeled.}
\label{fig:ins_doubleexp}
\end{figure}

\noindent Insult profiles $\vartheta$ varying in the $z_o$--$\theta_o$ (axial--azimuthal) plane are defined analytically with the expression \cite{Latorre2020b}

\begin{equation} \label{eq:doubleexp}
\vartheta(z_o,\theta_o) = \vartheta_{end} \, + \, \bigg( \vartheta_{apex} - \vartheta_{end} \bigg) \exp \left( - \left\vert \frac{z_o - z_{apex}}{z_{od}} \right\vert ^{\nu_z} \right)  \exp \left( - \left\vert \frac{\theta_o - \theta_{apex}}{\theta_{od}} \right\vert ^{\nu_\theta} \right),
\end{equation} \smallskip

\noindent where $z_o \in [0, l_o]$, $l_o$ is the initial axial length (15 mm), $\theta_o \in [0, 2\pi]$, $z_{od}$ and $\theta_{od}$ are the axial and circumferential characteristic widths of the insult region, respectively, $\nu_z$ and $\nu_\theta$ govern the softness of the boundaries in the axial and circumferential directions, respectively, and $\vartheta_{end}$ and $\vartheta_{apex}$ are values of the insult at the ends of the cylinder ($z_o = 0$, $l_o$) and the apex ($z_o = z_{apex}$, $\theta_o = \theta_{apex}$) of the profile, respectively. The profile is normalized to the interval $[0,1]$, with 1 indicating the maximum insult degree that varies depending on the insult type and severity. Example insult profiles and their corresponding remodeled vessels are shown in \autoref{fig:ins_doubleexp}, and parameter ranges are listed in \autoref{tab:params}.

\subsubsection{Randomly generated insult profiles}
\label{sec:methods:modeling_TAA_growth:random_insult_profiles}

\noindent In contrast to the definition above, we also model aneurysms initiated from insult profiles randomly distributed along the vessel wall to validate the DeepONet model against more natural (i.e., irregular) insults. These profiles are initially generated as ``latent'' (i.e., unobserved) Gaussian random fields (GRFs), then nonlinearly transformed and censored on $[0, 1]$ to match physically meaningful metrics prescribed by the user. On an unbounded space, the latent insult profiles are thus sampled according to

\begin{equation}
    \vartheta^\ast (z_o, \theta_o) \sim \mathcal{G} \big( \mu(z_o, \theta_o), \kappa(z_o, \theta_o, z_o', \theta_o') \big).
\end{equation} \smallskip

\begin{figure}[!htbp]
\centering
\includegraphics[width=0.95\textwidth]{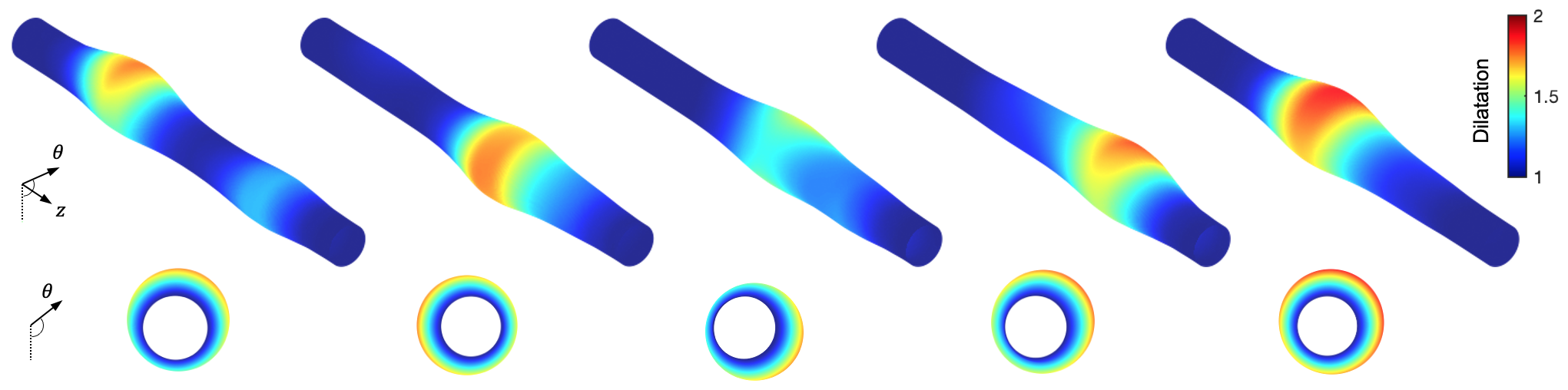}
\caption{3D and axial views of representative dilatation maps of TAAs initiated with prescribed insult profiles from mapped Gaussian random fields. Circumferential length scale $L_\theta = 1.5$~mm and axial length scale $L_z = 2.0$~mm with perturbation boundary softness $\epsilon = 0.6$. In all cases, $\varphi = 15$\% of the total area is perturbed, and the maximum insult applied is 60\% loss in elastic fiber integrity.}
\label{fig:ins_random}
\end{figure}

\noindent The mean $\mu(z_o, \theta_o)$ and covariance $\kappa(z_o, \theta_o, z_o', \theta_o')$ between two points $(z_o, \theta_o)$ and $(z_o', \theta_o')$ can be parameterized to control the overall propensity of insult ($\varphi$), the length scale of the insult(s) in the circumferential ($L_\theta$) and axial ($L_z$) directions, and the softness of the boundaries between normal and insult regions ($\epsilon$). To enforce periodicity in the circumferential direction, we define the covariance function as

\begin{equation}
\begin{aligned}
    \kappa(z_o, \theta_o, z_o', \theta_o') &= \varsigma^2 \exp \left( -\frac{1}{2} \left[ \left( \frac{D_\theta \left( \theta_o, \theta_o' \right)}{L_\theta} \right)^2 + \left( \frac{D_z \left( z_o, z_o' \right)}{L_z} \right)^2 \right] \right) \\
    \textrm{with} \quad D_\theta \left( \theta_o, \theta_o' \right) &= 2 r_o \sin{ \left( \frac{1}{2}\left| \theta_o - \theta_o' \right| \right) } \quad \text{and} \quad D_z \left( z_o, z_o' \right) = \left| z_o - z_o' \right|,
\end{aligned}
\end{equation} \smallskip

\noindent where $\varsigma^2$ is the overall variance of the GRF. To satisfy the prescribed insult propensity $\varphi$ and boundary softness $\epsilon$, the mean and variance of the GRF are defined as

\begin{equation}
\begin{aligned}
    \mu &= \frac{1}{2} - \frac{1}{\epsilon \sqrt{\pi}} \, \mathrm{erf}^{-1} \left( 1 - 2 \varphi \right) \exp \left( -\left[\mathrm{erf}^{-1} \left( 1 - 2 \varphi \right) \right]^2 \right) \\
    \textrm{and} \quad \varsigma^2 &= \frac{1}{2 \pi \epsilon^2} \exp \left( -2 \left[\mathrm{erf}^{-1} \left( 1 - 2 \varphi \right) \right]^2 \right),
\end{aligned}
\label{eqn:mean_and_variance}
\end{equation} \smallskip

\noindent where $\mathrm{erf}^{-1}( \cdot )$ denotes the inverse of the error function. Note that we choose $\mu$ to be constant herein with respect to $z_o$ and $\theta_o$. The insult propensity $\varphi$ corresponds to the fraction of $\vartheta^\ast$ values greater than 0.5, while $\epsilon$ corresponds to the slope of the cumulative distribution function (CDF) of $\vartheta^\ast$ at $\vartheta^\ast = 0.5$ (see \autoref{app:appendix_B:mean_and_variance} for the derivation of \autoref{eqn:mean_and_variance}). In practice, to ensure stability of the finite element simulations that follow, we constrain $\vartheta^\ast$ at the vessel boundaries to be low (e.g., two standard deviations below the mean), such that the censored insult profile values are zero at the boundaries. Note that $\vartheta^\ast$ is discretized on the finite element mesh of the vessel; thus, the nodal values $\vartheta_i^\ast$ jointly follow (and can thus be sampled from) a multivariate Gaussian distribution with mean vector $\bm{\mu} = \mu \mathbf{1}$ and covariance matrix $\bm{\Sigma}$, where $\Sigma_{ij} = \Sigma_{ji} = \kappa(z_{o,i}, \theta_{o,i}, z_{o,j}, \theta_{o,j})$. Partitioning the mesh into the set of interior nodes $a$ and the set of boundary nodes $b$, it is straightforward to condition the distribution of $\vartheta_a^\ast$ on the enforced value of $\vartheta_b^\ast$ using

\begin{equation}
\begin{aligned}
    \bm{\mu}_a' &= \bm{\mu}_a + \bm{\Sigma}_{ab} \bm{\Sigma}_{bb}^{-1} \left( \vartheta_b^\ast \bm{1} - \bm{\mu}_b \right) = \mu + \bm{\Sigma}_{ab} \bm{\Sigma}_{bb}^{-1} \left( \vartheta_b^\ast - \mu \right) \bm{1} \\
    \bm{\Sigma}_{aa}' &= \bm{\Sigma}_{aa} - \bm{\Sigma}_{ab} \bm{\Sigma}_{bb}^{-1} \bm{\Sigma}_{ba} \\
    \bm{\mu}_b' &= \vartheta_b^\ast \bm{1} \\
    \bm{\Sigma}_{ab}' &= \bm{0}, \quad \bm{\Sigma}_{ba}' = \bm{0}, \quad \bm{\Sigma}_{bb}' = \bm{0}
\end{aligned}
\end{equation} \smallskip

\noindent instead for sampling purposes. After $\vartheta_i^\ast$ is sampled from $\mathcal{N} \left( \bm{\mu}', \bm{\Sigma}' \right)$, we perform a CDF/inverse-CDF transformation, so that the overall distribution of $\vartheta^\ast$ values in each random instance of $\vartheta_i^\ast$ matches the desired $\mathcal{N} \left( \mu, \varsigma^2 \right)$ (see \autoref{app:appendix_B} for more details). Specifically,

\begin{equation}
    \left( \vartheta_i^\ast \right)' = \Phi^{-1} \left( F \left( \vartheta_i^\ast \right); \mu, \varsigma^2 \right),
\end{equation} \smallskip

\noindent where $F$ is the CDF of the generated random field values (approximated via kernel density estimation) and $\Phi^{-1}$ is the inverse CDF (i.e., quantile function) of the normal distribution with mean $\mu$ and variance $\varsigma^2$. Finally, the insult field values are censored using $\vartheta_i = \min \left( \max \left( \left( \vartheta_i^\ast \right)', 0 \right), 1 \right)$. The randomly generated insult profiles (i.e., loss of elastic fiber integrity or loss of mechanosensing) correspond in turn to random patterns of dilatation along the vessel (\autoref{fig:ins_random}), which are afterward used as input data for the DeepONet model to perform an inverse prediction of the insult profile. In \autoref{app:appendix_B:sensitivity}, we present a sensitivity analysis to demonstrate how $\varphi$, $\epsilon$, $L_\theta$, and $L_z$ jointly control the size, shape, and appearance of randomly generated insults with high precision.

\subsubsection{Insult contributors}
\label{subsubsec:instype}

\noindent Two types of contributors to TAAs are considered: loss of elastic fiber integrity (analogous to a loss of functional elastin) and dysfunctional mechanosensing (representing an impaired ability of cells to sense changes in intramural stress). Insult values ranging from mild to severe are simulated for each of these functions (\autoref{tab:params}).

\paragraph{Elastic fiber integrity} Functional elastin plays a crucial role in the elastic energy storage capability of the aorta (i.e., compliance and resilience), and genetic disorders such as Marfan syndrome associate with loss of elastic fiber integrity from fragmentation or degradation. This is modeled with a user-defined reduction in the baseline value of the material parameter $c^e$, chosen to achieve a loss in mechanical properties consistent with previous biaxial testing of aortas from \emph{Fbn1\textsuperscript{mgR/mgR}} mouse models for Marfan syndrome \cite{Lima2010, Bellini2016, Ramirez2018}.

\paragraph{Dysfunctional mechanosensing} Mechanical homeostasis of the aorta is regulated by intramural cells that sense the local microenvironment, primarily through integrin binding sites. When these cells cannot accurately detect deviations in stress from their homeostatic levels, whether through disruptions in extracellular matrix or impaired actomyosin activity, they tend to drive maladaptations within the aorta. This is represented with the parameter $\delta \in [0, 1]$, with 0 indicating perfect mechanosensing, and the modified expression for deviations in intramural stress $\Delta \sigma = ((1 - \delta)\sigma - \sigma_o)/\sigma_o$ \cite{Humphrey2015, Latorre2020b}.

\subsection{Finite element modeling}
\label{subsubsec:febio}

\subsubsection{Normotensive conditions}

\noindent We employ a 3D finite element model with $1 \times 20 \times 20$ mesh of 27-node quadratic hexahedral elements in the radial, circumferential (azimuthal), and axial directions. Displacement of the ends of the vessel is fixed in the axial direction, then the vessel is pressurized at normotensive conditions via a traction boundary condition on the inner surface and axially (pre-)stretched uniformly according to previous experimental data. The open-source software FEBio (\url{febio.org}) running a custom constrained mixture model plugin enables both G\&R computations and isochoric hyperelasticity computations in which G\&R is arrested \cite{Latorre2020a}. Simulations are performed in the following stages:

\begin{enumerate}
    \item Initialization: uniform pressurization and pre-stretch of the cylindrical vessel without insult to achieve the initial homeostatic state.
    \item G\&R: computation of growth and remodeling in response to the gradually applied insult, at fixed pressure.
    \item Hyperelasticity: post-G\&R computation of vessel deformation at normotensive diastolic and systolic pressures \cite{Bersi2016}.
\end{enumerate}

For analytically defined TAA cases, all combinations of insult profiles with variations in axial and circumferential location and extent were simulated for five levels of insult severity and for both insult types, yielding a total of 550 simulations at normotensive conditions. For randomly generated TAAs, 10 unique profiles sharing the same shape parameters are used to yield 100 cases.

\subsubsection{Superimposed hypertension}

\noindent Uncontrolled hypertension (elevated blood pressure) is a critical determining factor in TAA growth \cite{Bersi2016, Milewicz2019}. Thus, each insult type in \autoref{subsubsec:instype} is also simulated with superimposed hypertension, modeled by a gradual increase in the intraluminal pressure, concurrent with the prescribed insult. As a modification to the previously described simulation pipeline, hypertension cases are simulated as follows:

\begin{enumerate}
    \item Initialization: uniform pressurization and pre-stretch of the cylindrical vessel without insult to achieve the initial homeostatic state.
    \item G\&R: computation of growth and remodeling in response to the gradually applied insult, concurrent with gradual increases in pressure by 33\%.
    \item Hyperelasticity: post-G\&R computation of vessel deformation at hypertensive diastolic and systolic pressures \cite{Bersi2016}.
\end{enumerate}

Cases of hypertension highlight how even a mild insult degree with modest dilatation under normotensive conditions can be exacerbated by increased pressure, yielding a maximum dilatation comparable to that produced by a severe insult under normotensive conditions (\autoref{tab:params}). 550 additional simulations with hypertensive conditions are simulated for analytically defined profiles. Hypertensive conditions were not considered for randomly generated insults.

\begin{table}[!htbp]
\centering
\caption{G\&R and mechanobiological insult profile parameters for normotensive and hypertensive simulations with analytically defined insults (\autoref{eq:doubleexp}). To avoid boundary effects, $z_{apex} = 7.5$ mm only for $z_{od} = 4.0$ mm. For randomly generated profiles, loss of elastic fiber integrity is varied from 6\% -- 60\%, and dysfunctional mechanosensing ranges from 2.5\% -- 25\% at normotensive conditions.}
\setlength{\tabcolsep}{20pt}
\resizebox{\textwidth}{!}{
\begin{tabular}{ l c c c }
\toprule
Parameter & Variable & \multicolumn{2}{c}{Value} \\
\hline
Axial characteristic width (mm) & $z_{od}$ & \multicolumn{2}{c}{2.0, 3.0, 4.0} \\
Circumferential characteristic width (deg) & $\theta_{od}$ & \multicolumn{2}{c}{$20$, $100$, $180$, $260$, $360$} \\
Axial placement (mm) & $z_{apex}$ & \multicolumn{2}{c}{6.0, 7.5, 9.0} \\
Circumferential placement (deg) & $\theta_{apex}$ & \multicolumn{2}{c}{$0$, $90$, $180$, $270$} \\
\hline
& & Normotensive & Hypertensive (33\%) \\
\hline
Loss of elastic fiber integrity & & 5.95\% -- 59.5\% & 4.75\% -- 47.5\% \\
Dysfunctional mechanosensing & & 1.84\% -- 18.4\% & 1.08\% -- 10.8\% \\
G\&R pressure (mmHg) \cite{Latorre2020b} & & 105 & $105 \rightarrow 140$ \\
Diastolic pressure (mmHg) \cite{Bersi2016} & & 99 & 129 \\
Systolic pressure (mmHg) \cite{Bersi2016} & & 121 & 172 \\
\bottomrule
\end{tabular}}
\label{tab:params}
\end{table}

\subsubsection{Post-processing training and testing data}
\begin{figure}[!htbp]
\centering
\includegraphics[width=0.8\textwidth]{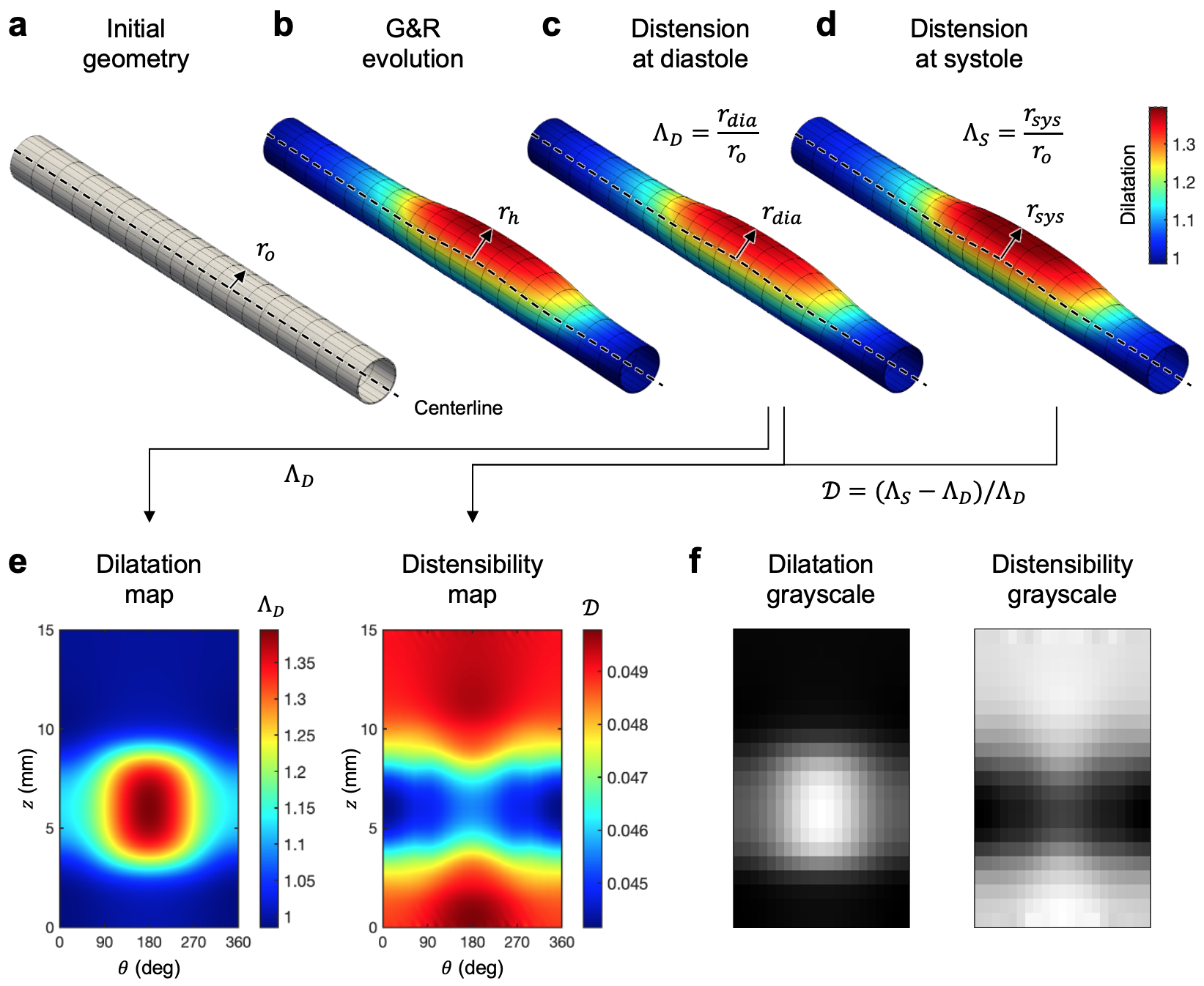}
\caption{Schematic for the generation of synthetic training data for TAAs arising from localized mechanobiological insults. (a) The initial vessel geometry with inner radius $r_o$ and prescribed insult is evolved over multiple load steps to compute (b) the final post-G\&R geometry, in which the homeostatic inner radius $r_h$ can be calculated with respect to the updated centerline. The luminal pressure is then adjusted to (c) diastolic and (d) systolic conditions to evaluate the local distension at diastole $\Lambda_D$ and systole $\Lambda_S$, respectively. (e) The distensibility is then computed with the relation $\mathcal{D} = (\Lambda_S - \Lambda_D)/\Lambda_D$. Together, these steps yield the dilatation $\Lambda_D$ and distensibility $\mathcal{D}$ maps (shown in the flattened $z$--$\theta$ plane). (f) Finally, the maps are converted into nondimensionalized grayscale intensity maps and contrast enhanced. For illustrative purposes, this procedure is shown for an analytically defined insult (\autoref{eq:doubleexp}) under normotensive conditions only.}
\label{fig:postproc}
\end{figure}

\noindent In the Initialization stage, the straight cylindrical vessel with inner radius $r_o$ is uniformly pressurized and pre-stretched according to normotensive \emph{in vivo} conditions in one load step while maintaining the same geometry (\autoref{fig:postproc}a). In the G\&R stage, an insult (analytically defined or randomly generated) is gradually applied over 10 subsequent load steps to compute the evolved post-G\&R geometry (\autoref{fig:postproc}b) with spatially varying inner radius $r_h$. The G\&R evolution is then arrested, and in the Hyperelasticity stage two additional load steps are simulated in which the luminal pressure is adjusted to diastolic and systolic values \cite{Bersi2016} to evaluate the local distension at diastole $\Lambda_D = r_{dia}/r_o$ and systole $\Lambda_S = r_{sys}/r_o$, where $r_{dia}$ and $r_{sys}$ are the inner radius at diastolic and systolic pressure, respectively (\autoref{fig:postproc}c,d).

We train the DeepONet surrogate model using two quantities of interest from the finite element-based synthetic data: dilatation (i.e., aneurysmal enlargement) and distensibility (i.e., cyclic strains). For dilatation, we utilize a map of $\Lambda_D$ for each mesh node along the inner surface of the geometry. For distensibility, in contrast to the diameter- and pressure-based ``distensibility'' defined in previous literature \cite{Bersi2016}, we consider the actual distensibility $\mathcal{D} = (\Lambda_S - \Lambda_D)/\Lambda_D$ for all nodes on the inner surface. These dilatation and distensibility maps are provided to the neural network in the form of $z$--$\theta$ maps (\autoref{fig:postproc}e) as well as nondimensionalized grayscale images (\autoref{fig:postproc}f). Approximately 10\% of the total number of datasets are used to validate the trained model, and the remaining are used for training.

\subsection{DeepONet}
\label{subsec:deeponet}



\noindent In this section, we describe the architecture of the surrogate model developed within the framework of DeepONet to predict mechanobiological insult profiles. The conventional unstacked DeepONet architecture consists of two deep neural networks (DNNs): one encodes the input function at fixed sensor points (branch net) while the other accounts for the locations of the output function (trunk net). The branch network input is customizable in a generalized setting and can take the shape of the physical domain, the initial or boundary conditions, constant or variable coefficients, source terms, and so on, as long as the input function is discretized at sensor locations $n_{sen}$. For a regularly spaced discretization of the input function, a convolutional neural network (CNN) can be used as the branch net, while for a sparse representation, one may also consider a feed-forward neural network (FNN), or even a recurrent neural network (RNN) for sequential data. A standard practice is to use an FNN in the trunk network to take into account the low dimensions of the evaluation points. The mathematical foundation of DeepONet is provided in \autoref{app:appendix_c}, and the interested reader is referred to \cite{lu2021learning} for additional details.

The cases considered in this work consist of mechanobiologically equilibrated TAAs developed from an initially cylindrical vessel caused by loss of elastic fiber integrity or compromised mechanosensing. We generate cases under normotensive and hypertensive conditions and consider dilatation and distensibility maps to train the DeepONet. We consider two different network architectures for the branch network, FNN and CNN, with the aim to identify a preferred network architecture. With an FNN in the branch network, we show that even with sparse dilatation and distensibility information, the DeepONet predicts the insult profile with high accuracy. Subsequently, we use a CNN to approximate the input function with grayscale images of the dilatation and distensibility. Both approaches are judged based on two empirical observations:

\begin{itemize}
    \item the ability of the models to generalize unseen test cases, and
    \item the ability of these models to generalize noisy input functions.
\end{itemize}

\subsubsection{FNN-based architectures}
\label{subsubsec:FNN_branch}

\noindent The DeepONet in this work considers multiple branch networks to account for the dilatation map at diastole ($\Lambda_S$) and the distensibility map ($\mathcal{D}$), as well as whether the loading condition is normotensive or hypertensive. The five branch networks $\mathbf{U}^i$ for $i = 1,2,\ldots,5$ are formulated as FNNs with the following description:

\begin{itemize}
    \item $\mathbf{U}^1$: value of $\Lambda_D$ at $n_{sen}$.
    \item $\mathbf{U}^2$: location of the maximum value in $\Lambda_D$.
    \item $\mathbf{U}^3$: value of $\mathcal{D}$ at $n_{sen}$.
    \item $\mathbf{U}^4$: location of the minimum value in $\mathcal{D} = (\Lambda_S - \Lambda_D)/\Lambda_D$.
    \item $\mathbf{U}^5$: binary network to account for hypertension: \emph{normotensive: 0, hypertensive: 1}.
\end{itemize}

\noindent We perform two experiments to decide the optimal number of sensors for $\mathbf{U}^1$ and $\mathbf{U}^3$, detailed below.

\bigskip
\noindent \textbf{Experiment 1:} $n_{sen} = 5\times5$ single-spaced sensor locations.

\begin{figure}[!ht]
\centering
\includegraphics[width=0.75\textwidth]{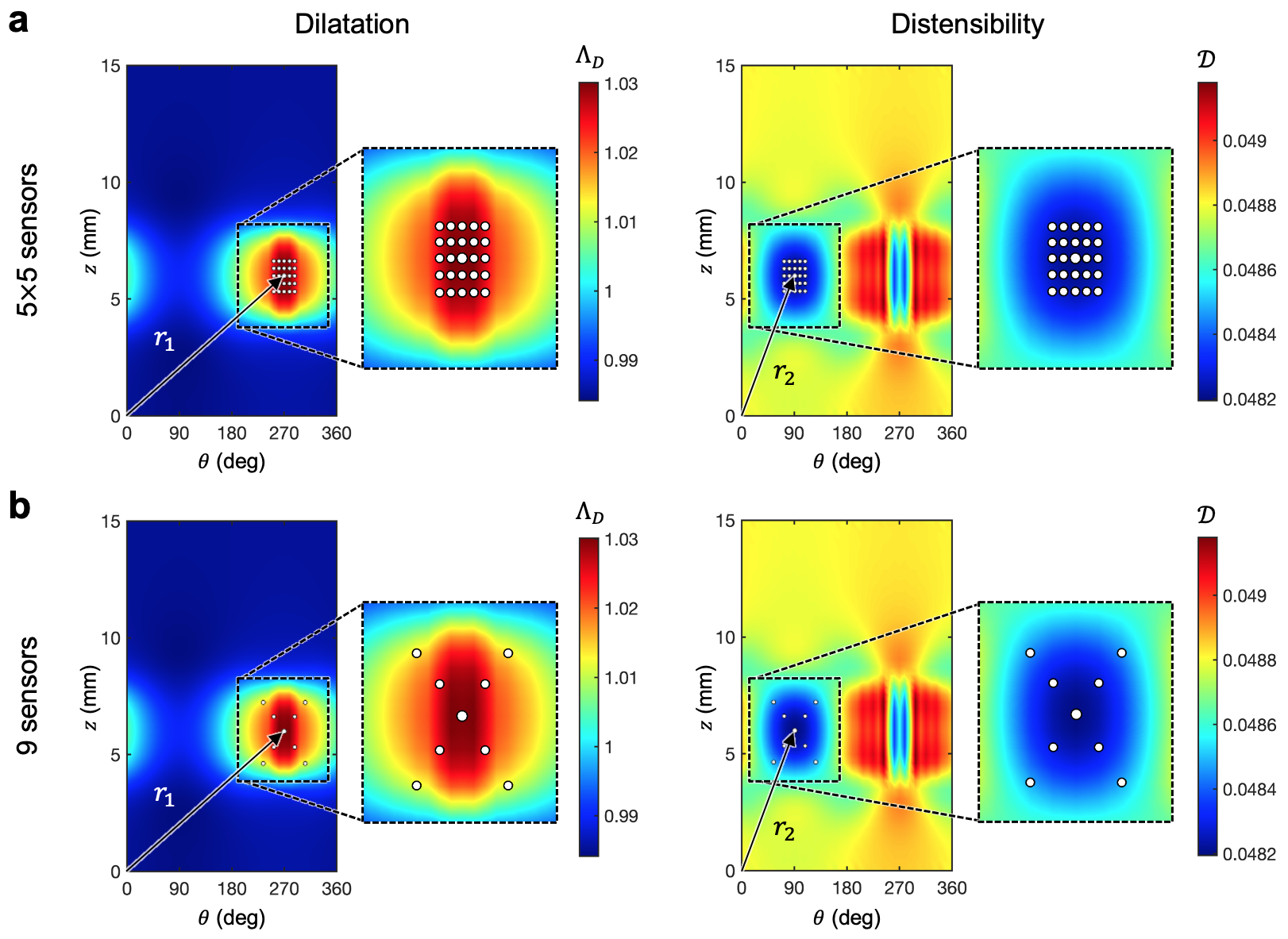}
\caption{Depictions of the sensor locations for \autoref{subsubsec:FNN_branch} within a representative TAA case (10.1\% loss of mechanosensing, $z_{od} = 2$ mm, $\theta_{od} = 20^\circ$, $z_{apex} = 6$ mm, $\theta_{apex} = 270^\circ$). (a) $n_{sen} = 5 \times 5$ single-spaced sensor locations on $\Lambda_D$ and $\mathcal{D}$ for Experiment 1. (b) Double-spaced sensor locations $n_{sen} = 9$ for Experiment 2. The scalar distances $r_1$ and $r_2$ are computed from the origin to the maximum dilatation location and the minimum distensibility location, respectively.}
\label{fig:sensorlocs}
\end{figure}

\noindent In this experiment, a $5\times5$ lattice of points is employed for $\mathbf{U}^1$ and $\mathbf{U}^3$, centered at the location of maximum dilatation $\mathbf{U}^2$ and the location of minimum distensibility $\mathbf{U}^4$. The locations of the sensors are selected such that the input data is constrained within a fairly localized neighborhood around the local extremum, dictated by the spacing of the nodes in the finite element mesh (in other words, the sensor locations correspond to a uniform $5\times5$ grid of adjacent single-spaced nodes). The arrangement of sensor points is shown in \autoref{fig:sensorlocs}a.

\bigskip
\noindent \textbf{Experiment 2:} $n_{sen} = 9$ double-spaced sensor locations.

\noindent In this experiment, only $9$ sensor locations are employed for for $\mathbf{U}^1$ and $\mathbf{U}^3$; however, locations are distributed with double the spacing in the axial and azimuthal directions compared to the single-spaced $5\times5$ locations, encompassing a neighborhood roughly 4 times the size of that in Experiment 1. As above, the sensors for $\mathbf U^2$ and $\mathbf U^4$ are centered at the locations of maximum dilatation and minimum distensibility, respectively (\autoref{fig:sensorlocs}b).

\begin{figure}[!ht]
\centering
\includegraphics[width=\textwidth]{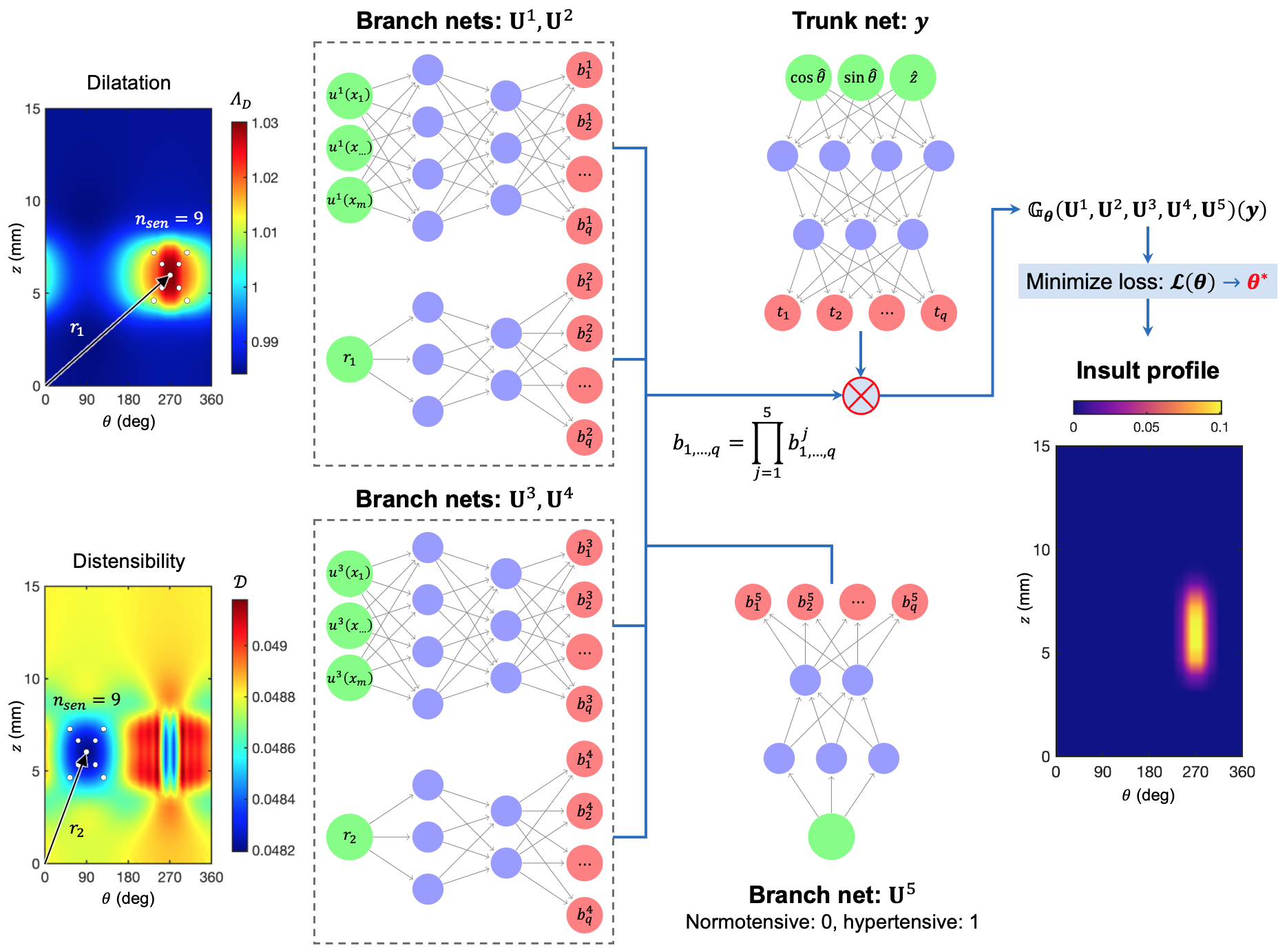}
\caption{Schematic representation of the DeepONet for \autoref{subsubsec:FNN_branch}. The branch network takes as input the dilatation $\Lambda_D$ and distensibility $\mathcal{D}$ at the sensor locations $n_{sen}$ (branch net: $\mathbf{U}^1$ and $\mathbf{U}^3$, respectively) and the locations of maximum dilatation and minimum distensibility (branch net: $\mathbf{U}^2$ and $\mathbf{U}^4$, respectively). Additionally, blood pressure information is fed to the network using a branch net: $\mathbf{U}^5$ as normotensive (input $= 0$) or hypertensive (input $= 1$). Each branch net returns features embedding $[b^i_1, b^i_2, \ldots, b^i_q]^\mathrm{T} \in \mathbb{R}^q$, $i={1,2,\ldots,5}$ as output, and the dot product of these outputs is computed to obtain $[b_1, b_2, \ldots, b_q]^\mathrm{T}$. The trunk net takes the continuous coordinates $\yb = (\cos\hat \theta, \sin\hat\theta, \hat z)$ as inputs and outputs a features embedding $[t_1, t_2, \ldots, t_q]^\mathrm{T} \in \mathbb{R}^q$. The features embeddings of the branch and trunk networks are merged via an element-wise dot product to output the solution operator $\mathbb G_{\bm{\theta}}$ with learnable parameters $\bm \theta$. Minimization of the loss function $\mathcal L$ determines best-fit parameters $\bm \theta^\ast$ that enable estimation of the insult profile.}
\label{fig:DeepONet1}
\end{figure}

The trunk network (FNN) considers the locations of evaluation in a cylindrical coordinate system (i.e., $\bm y = \{\yb_1,\yb_2,\ldots,\yb_p\} = \{(\hat \theta_1,\hat z_1),(\hat \theta_2, \hat z_2), \ldots, (\hat \theta_p,\hat z_p)\}$). Encapsulating previous knowledge into the architecture of DeepONet often improves the generalization of the output. Here, we replace the $\hat\theta$ component of the trunk net with appropriate basis functions in Cartesian coordinates. Thus, the trunk net input is modified as $\yb_i = (\cos\hat \theta_i, \sin\hat\theta_i, \hat z_i)$. A schematic representation of the framework is shown in \autoref{fig:DeepONet1}.

\subsubsection{CNN-based architecture}
\label{subsubsec:CNN_branch}


\noindent In this section, we discuss the DeepONet architecture considering grayscale images, converted from the $\Lambda_D$ and $\mathcal{D}$ maps and contrast enhanced, as input functions for the branch network. This approach considers 3 branch networks with the following descriptions:

\begin{itemize}
    \item $\mathbf{U}^1$: grayscale image for $\Lambda_D$ using a CNN.
    \item $\mathbf{U}^2$: grayscale image for $\mathcal{D}$ using a CNN.
    \item $\mathbf{U}^3$: binary network to account for hypertension using an FNN.
\end{itemize}

\begin{figure}[!ht]
\centering
\includegraphics[width=\textwidth]{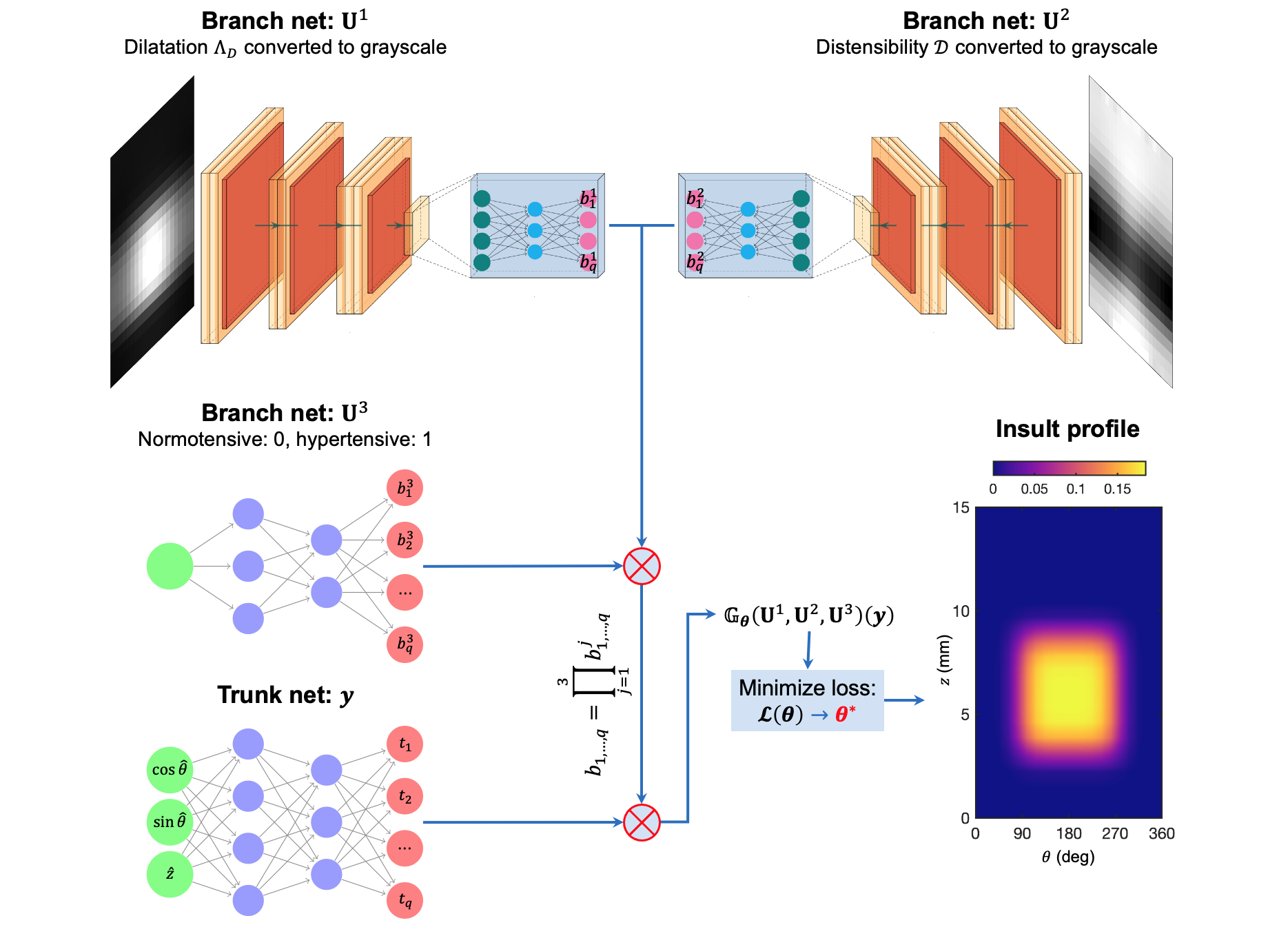}
\caption{Schematic representation of the DeepONet for \autoref{subsubsec:CNN_branch}. Each branch net is a CNN that takes as inputs the grayscale images of dilatation $\Lambda_D$ and distensibility $\mathcal{D}$ (branch nets: $\mathbf{U}^1$ and $\mathbf{U}^2$, respectively). Blood pressure information is fed to the FNN using a branch net: $\mathbf{U}^3$ as normotensive (input $= 0$) or hypertensive case (input $= 1$). The outputs of the branch and trunk networks are merged into the solution operator $\mathbb G_{\bm{\theta}}$, and minimization of the loss function $\mathcal L$ enables estimation of the insult profile.
}
\label{fig:DeepONet2}
\end{figure}

\noindent Note that, in contrast to the sparse sensor point arrays, we use full-field images for $\mathbf{U}^1$ and $\mathbf{U}^2$, which have a resolution of $21\times 20$ pixels. A schematic representation of the DeepONet implemented with this framework is shown in \autoref{fig:DeepONet2}.

\section{Results}
\label{sec:results}

\subsection{Dilatation and distensibility in TAA}

\noindent Locally applied insults in elastic fiber integrity and mechanosensing both lead to the development of dilatations, as well as aneurysms (defined as a 1.5-fold increase in normalized diameter from baseline) in cases of severe insults of elastic fiber integrity. In cases with less circumferential involvement of insult ($\theta_{od} < 180^\circ$), the unaffected regions of the vessel help to attenuate dilatation within the insult area. Additionally, each factor is modeled in combination with superimposed hypertension (elevated blood pressure), emphasizing how even a mild insult with modest dilatation in normotensive conditions can be exacerbated by the presence of additional risk factors. In hypertensive cases, the dilatation also increases significantly, including in regions where there is no insult applied.

Overall, increased dilatation associated with mechanobiological insults correlate well with decreases in distensibility. However, there are distinct differences between distensibility maps resulting from loss of elastic fiber integrity and those resulting from dysfunctional mechanosensing. Whereas the location of maximum dilatation and minimum distensibility colocalize in elastic fiber integrity loss, the maximum dilatation in mechanosensing loss occurs on the opposite side of the vessel from the minimum distensibility, although the distensibility also decreases at the location of maximum dilatation. These findings confirm the need to evaluate both metrics for robust training of the DeepONet in all cases, especially in experiments considering multiple types of insults.

\subsection{Performance of the surrogate models}

\noindent The effectiveness of the developed surrogate models is demonstrated through several experiments, put forward in this section. To evaluate performance, we compute the $L_2$ relative error of predictions, and we report its mean and standard deviation based on five independent training trials. In all cases presented here, the DeepONet is trained using a combination of Adam \cite{kingma2014adam} and L-BFGS optimizers \cite{liu1989limited}. The implementation is carried out using the \texttt{TensorFlow} framework \cite{abadi2015tensorflow}. Throughout all examples, we initialize the weights and biases of the DeepONet using Xavier initialization. The experiments carried out in this work are listed in \autoref{table:cases}.

\begin{table}[ht!]
\caption{Descriptions of the datasets considered for the experiments carried out in this work. Insult profiles are either analytically defined or randomly generated losses of elastic fiber integrity or mechanosensing, and pressure conditions are either normo- or hypertensive, listed in parentheses.}
\footnotesize
\centering
\begin{tabular}{c l}
\toprule
Case \# & Description \\
\hline
Case 1 & Analytically defined elastic fiber integrity (normotensive) \\
Case 2 & Analytically defined mechanosensing (normotensive) \\
Case 3 & Analytically defined elastic fiber integrity or mechanosensing (normotensive) \\
Case 4 & Analytically defined mechanosensing (normotensive \& hypertensive) \\
Case 5 & Analytically defined elastic fiber integrity or mechanosensing (normotensive \& hypertensive) \\
Case 6 & Randomly generated elastic fiber integrity or mechanosensing (normotensive) \\
\bottomrule
\end{tabular}
\label{table:cases}
\end{table}

\begin{figure}[!ht]
\centering
\includegraphics[width=0.8\textwidth]{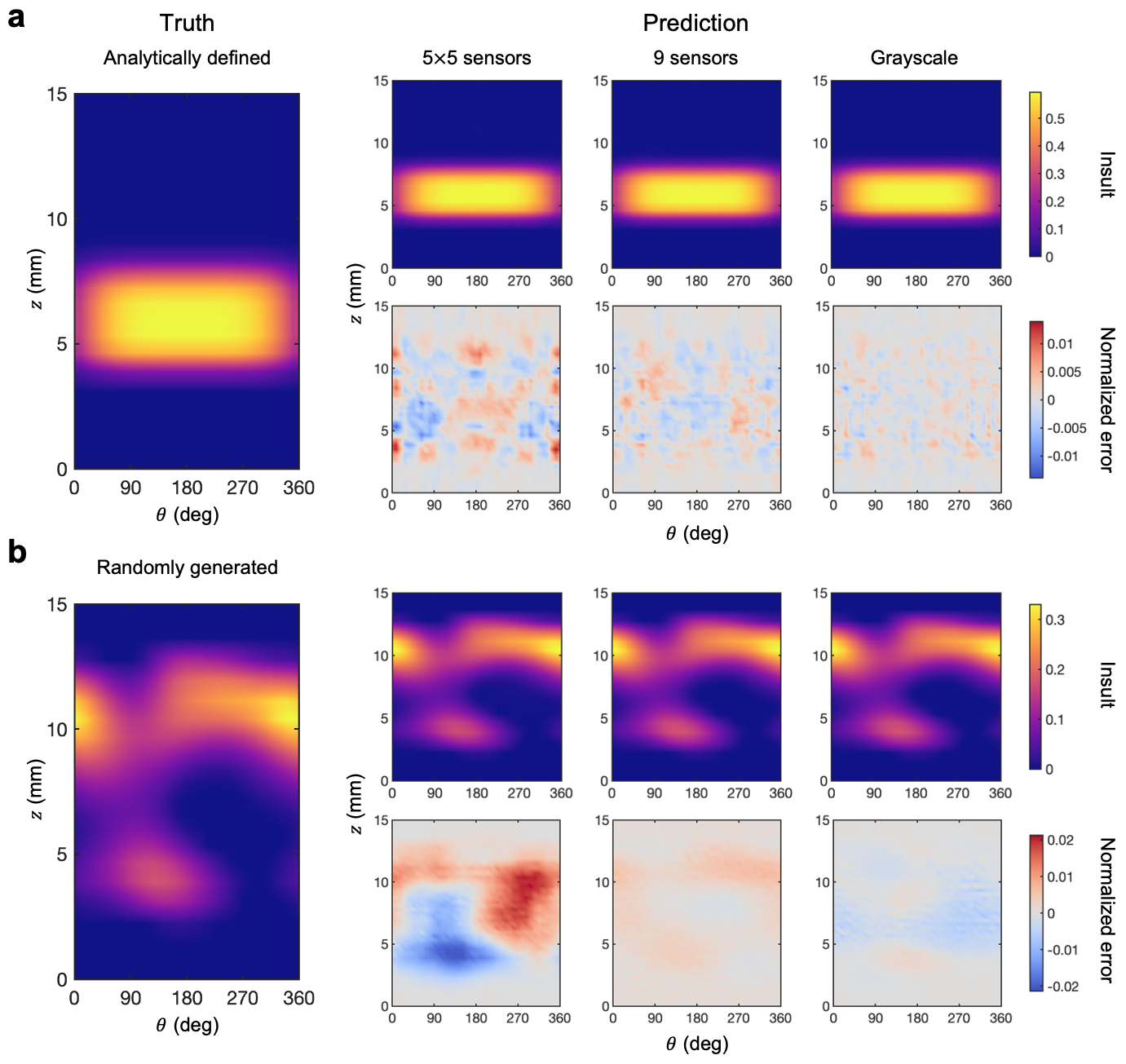}
\caption{Performance of the surrogate models for the prediction of analytically defined and randomly generated insult profiles. Axial--azimuthal views of representative true and predicted (a) analytically defined and (b) randomly generated insult profiles are predicted using $5\times5$ sensors, $9$ sensors, and full-field grayscale maps. The insult prediction error, normalized by the maximum insult value over the axial--azimuthal domain, is shown for each architecture.}
\label{fig:summary}
\end{figure}

The experiments establish the accuracy of the three surrogate models when trained with data generated for analytically defined insult profiles (Cases 1--5) and for randomly generated insult profiles (Case 6). In \autoref{fig:summary}, we present the performance of the three surrogate models for representative analytically defined and randomly generated insults in elastic fiber integrity (Cases 1 \& 6). For each architecture, we compute an error between the true and predicted insult profiles by normalizing the absolute error by the maximum insult value throughout the $z$--$\theta$ plane. While each DeepONet architecture is able to predict the insult profile within 5\% error, the design with $5\times5$ sensors tends to exhibit the greatest prediction errors, whereas the design based on grayscale images provides the most accurate prediction. This discrepancy is most clearly observed in \autoref{fig:summary}b.

In \autoref{table:R4_norm}, we show the relative $L_2$ error of the surrogate models, averaged over the $z$--$\theta$ domain, for the testing dataset compromised of mechanosensing insults under normotensive conditions (Case 2). Additionally, we show the total number of trainable (i.e., learnable) parameters for each surrogate model, which are the weights and biases of the networks optimized using back-propagation during the training process. In general, the total number of parameters learned during the training process for a CNN is less than that for an FNN. CNNs are very effective in reducing the number of parameters without losing the quality of models, as the learnable filters of the network encode the information to reduce the high-dimensional input data (here, grayscale images) to a lower latent dimension. This is also seen in our results, which show that the network with full-field grayscale images requires the lowest number of parameters while simultaneously achieving the highest predictive accuracy. Finally, we evaluate the performance of the surrogate model trained with grayscale images on a noisy dataset that consists of the original testing dataset with $5\%$ added uncorrelated Gaussian noise. Overall, among the surrogate models considered in this Case, the CNN-based design again demonstrates improved computational efficiency, prediction accuracy, and robustness against noise.

\begin{table}[ht!]
\footnotesize
\caption{Relative $L_2$ error and the number of trainable parameters of the three surrogate models for $N = 545$ training data and $N_* = 45$ testing data consisting of compromised mechanosensing under normotensive conditions. The noise is added in the inputs of the \emph{testing} dataset.}
\centering
\begin{tabular}{l c c c}
\toprule
\multirow{2}{*}{Method} & \multirow{2}{*}{\# parameters} &\multicolumn{2}{c}{Relative $L_2$ error} \\ \cmidrule(l){3-4} 
&  & Testing data & $+5\%$ noise \\
\toprule
$5\times5$ sensors & $234800$ & $3.906\pm 0.004\%$ & $44.27 \pm 0.008\%$\\
$9$ sensors & $230784$ & $2.712\pm 0.007\%$ & $32.52 \pm 0.009\%$ \\
Grayscale images & $141180$ & $2.340 \pm 0.010\%$ & $8.2\pm 0.020\%$\\
\bottomrule
\end{tabular}
\label{table:R4_norm}
\end{table}

\begin{figure}[!ht]
\centering
\includegraphics[width=\textwidth]{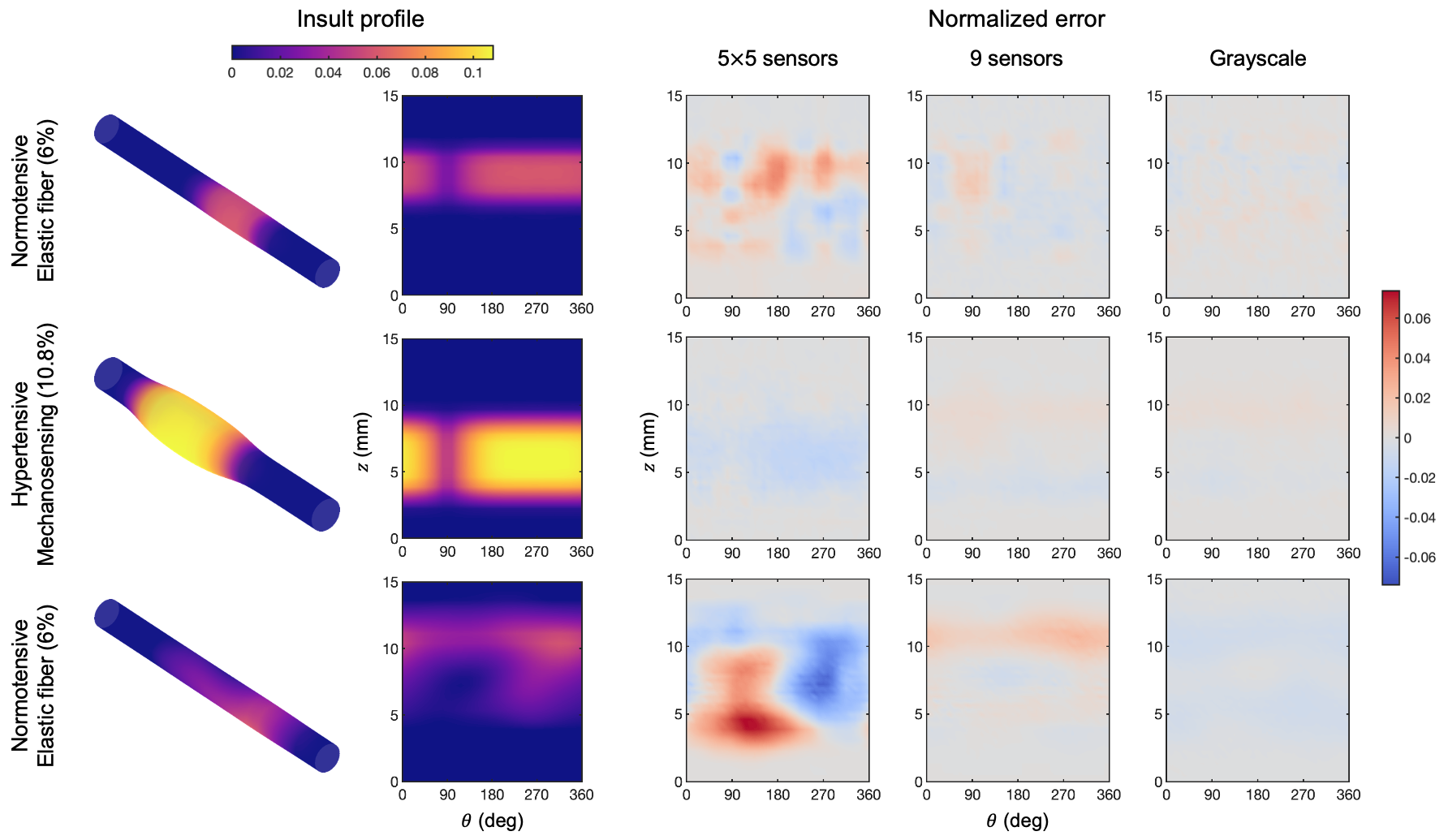}
\caption{Error plots for three representative cases of testing samples from the indicated experiments. Columns 1 and 2 show the 3D geometry and the unfolded axial--azimuthal view of the aorta, respectively, colored by the insult profile (ground truth). Columns 3--5 depict the prediction error of Experiment 1 (using $5\times5$ sensors), Experiment 2 (using $9$ sensors), and using grayscale images to train the network, respectively. Error is normalized with respect to the maximum insult value of each case. The top row corresponds to a case of analytically defined normotensive elastic fiber integrity insult (Case 1), the middle row shows an analytically defined hypertensive mechanosensing insult (Case 4), and the bottom row represents a randomly generated normotensive elastic fiber integrity insult (Case 6).}
\label{fig:error_plots}
\end{figure}

In \autoref{table:all_errors}, we report the relative $L_2$ errors of the three models, along with the robustness of each model to $5\%$ added uncorrelated Gaussian noise. Error plots for three representative test cases are shown in \autoref{fig:error_plots}. Columns 1 and 2 show the 3D geometry and the unfolded axial--azimuthal view of the aorta, respectively, colored by the insult profile (ground truth). Columns 3--5 depict the prediction errors associated with using $5\times5$ sensors, $9$ sensors, or grayscale images to train the network, respectively. Error is normalized with respect to the maximum insult value of each case. These findings further reinforce the improvements of the CNN-based approach in a variety of testing scenarios varying by insult type, pressure conditions, and insult profile generation method.

\begin{table}[ht!]
\caption{Relative $L_2$ error of the three surrogate models for with $N$ training data and $N_*$ testing data for the cases listed in \autoref{table:cases}. The noise is added in the grayscale full-field image inputs of the \emph{testing} dataset.}
\footnotesize
\centering
\begin{tabular}{c c c c c c c}
\toprule
\multirow{2}{*}{Case \#} & \multirow{2}{*}{$N$} & \multirow{2}{*}{$N_*$} & \multicolumn{4}{c}{Relative $L_2$ error} \\ 
\cmidrule(l){4-7} & & & $5\times5$ sensors & $9$ sensors & Grayscale images & $+5\%$ noise\\
\toprule
Case 1 & $545$ & $45$ & $4.79\pm 0.008\%$ & $4.29\pm 0.008\%$ & $3.458 \pm 0.011\%$ & $7.80\pm 0.03\%$\\
Case 2 & $545$ & $45$ & $3.91\pm 0.004\%$ & $2.71\pm 0.007\%$ & $2.340 \pm 0.010\%$ & $8.20\pm 0.020\%$\\
Case 3 & $500$ & $90$ & $7.10\pm 0.001\%$ & $6.75\pm 0.003\%$ & $5.948\pm 0.006\%$ & $10.69 \pm 0.015\%$\\
Case 4 & $500$ & $90$ & $2.88\pm 0.005\%$ & $2.62\pm 0.001\%$ & $2.280 \pm 0.005\%$ & $7.30 \pm 0.01\%$\\
Case 5 & $720$ & $180$ & $3.65\pm 0.005\%$ & $2.65\pm 0.003\%$ & $2.540 \pm 0.018\%$ & $7.18 \pm 0.012\%$\\
Case 6 & $90$ & $10$ & $7.74\pm 0.011\%$ & $7.20\pm 0.020\%$ & $2.292 \pm 0.002\%$ & $15.96 \pm 0.029\%$\\
\bottomrule
\end{tabular}
\label{table:all_errors}
\end{table}

\section{Discussion}
\label{sec:discussion}

\subsection{Insult profile prediction with sensor point- and image-based approaches}

\begin{figure}[!ht]
\centering
\includegraphics[width=\textwidth]{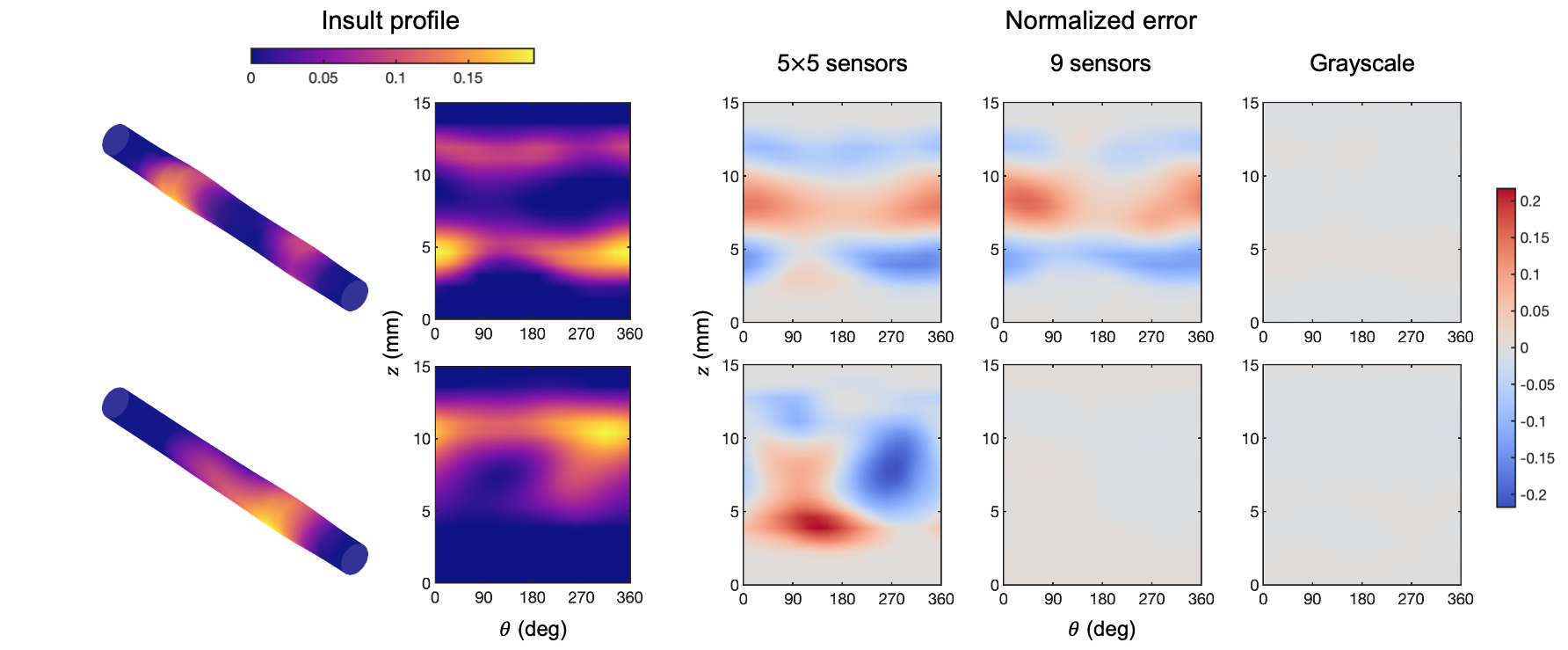}
\caption{Error plots for two representative cases of loss of elastic fiber integrity under normotensive conditions for randomly generated insult profiles (Case 6). Columns 1 and 2 show the 3D geometry and unfolded view of the aorta with aneurysm (ground truth), respectively, and columns 3--5 depict the normalized prediction error of Experiment 1 (using $5\times5$ sensors), Experiment 2 (using $9$ sensors), and using grayscale images to train the network, respectively.}
\label{fig:Random-R1+R4-1_error}
\end{figure}

\noindent This study utilizes a constrained mixture model for arterial growth and remodeling integrated with a DeepONet to predict factors contributing to TAA. The main challenge in applying deep learning based frameworks to predict TAA enlargement is the availability of limited information, both in terms of number of samples as well as information per sample. Additionally, due to the high complexity of the mechanobiology in patient-specific predictions of TAA (e.g., age, hypertension, diabetes, prescribed medications), making accurate predictions with even high-fidelity physical models remains a challenge. To enable accurate predictions, we have proposed and evaluated three frameworks of DeepONet that are distinguished based on the information available for training the network. The results shown in \autoref{table:all_errors} indicate that the proposed frameworks indeed provide effective predictions for both analytically defined and randomly generated insult profiles. Our observations are summarized as follows:

\begin{enumerate}
    \item It has been shown that aortic geometry alone is not sufficient to predict TAA progression \cite{Ostberg2022}. Many have thus sought to incorporate additional information such as biomechanical properties and patient-level variables to improve predictive capability. We find that accurate prediction of the insult profile can be achieved with the inclusion of biomechanical properties along with dilatation and distensibility fields generated from measurements of diastolic and systolic phases of the cardiac cycle, similar to the approaches used by other groups \cite{Liu2021, LindquistLiljeqvist2021}.
    \item Predicting insult profiles with information at $5\times5$ sensor locations is sufficiently accurate for cases with $\theta_{od}<260^{\circ}$ and relatively small $z_{od}$. However, with a wider $\theta_{od}$ and broader $z_{od}$, limited information within a single-spaced neighborhood around maximum dilatation and minimum distensibility is insufficient; hence, for such cases, the relative error increases.
    \item Overcoming the limitations of $5\times5$ sensor locations, the double-spaced arrangement of $9$ sensors more accurately estimates the insult profile within a wider neighborhood compared to that of the $5\times5$ sensor array. This observation is in line with the reduced prediction errors reported in \autoref{table:all_errors} (fourth and fifth column), which depicts this improvement in accuracy. This suggests that the range of variation within the network inputs captured by the sensor point domain, rather than the absolute number of sensors, is a greater determining factor in the predictive capability of the network.
    \item FNNs have proven to work well with limited information and also be robust to noisy testing inputs. However, in the case of extremely sparse information, the network fails to generalize well for noisy inputs and reports a relative $L_2$ error of $44.27 \pm 0.008\%$ and $32.52 \pm 0.009\%$ for $5\times5$ and $9$ sensors, respectively, when tested with $5\%$ Gaussian noise added to the testing inputs of Case 2 (compromised mechanosensing) (last column in \autoref{table:R4_norm}).
    \item The robustness of the different network architectures is evident in cases of randomly generated insult profiles. As seen in the third and the fourth columns of \autoref{fig:Random-R1+R4-1_error}, networks trained on sparse sensor point information often fail to achieve accurate predictions if there is more than one location of localized dilatation or if the dilatation boundary is irregular. In these cases, we observe that the model with $9$ double-spaced sensor locations achieves better prediction accuracy than the $5\times5$ single-spaced sensor locations.
    \item The caveats of the FNN-based frameworks (\autoref{subsubsec:FNN_branch}) motivate the development of a CNN-based framework, which takes as inputs grayscale images of dilatation and distensibility to more accurately predict the insult profile. The predictive accuracy of the framework proposed in \autoref{subsubsec:CNN_branch} is clearly depicted in \autoref{table:all_errors} (sixth column) and \autoref{fig:error_plots}. Additionally, the model is more noise-tolerant than the sparse sensor-based frameworks (last column of \autoref{table:all_errors}).
    \item The computational cost of a network is directly related to the number of trainable parameters. Neural networks employ back-propagation algorithms to tune the network parameters, while trying to reflect the best-fit solution to the training data. Therefore, an FNN trained on real-life images is often computationally expensive. The choice of a CNN to train on grayscale images benefits the model not only in terms of predictive accuracy but also computational efficiency, as this framework requires fewer learnable parameters compared to the models employing FNNs (second column of \autoref{table:R4_norm}).
    \item All of the frameworks proposed are capable of handling patient-specific datasets beyond dilatation and distensibility. In this work, we have demonstrated that an additional FNN in the branch net is sufficient to handle information like hypertension. This framework can be easily extended to handle further information, such as preexisting genetic or pharmacological conditions, without any additional cost of training the network.
\end{enumerate}

\subsection{Limitations}

\noindent There are also restrictions to the frameworks proposed and experiments selected in this study, which will be addressed in future work. The proposed framework is capable of predicting the insult profile for a given set of dilatation and distensibility information (that is, for a given time). This capability represents a scientifically important inverse problem, a class of problems that are typically ill-posed and yet were tractable herein. Nevertheless, even though the three surrogate models are capable of handling patient-specific information, they are not yet capable of determining real-time TAA evolution, which is a significant unmet need in the assessment of TAA for diagnosis, prognosis, and follow-up, and also in the determination of rupture risk. The study has been carried out on synthetic data based on finite element simulations, which are available in abundance. Future training of the model on limited clinical information will be challenging in terms of the number of data points available. In such situations, transfer learning may need to be integrated with domain adaptation models to leverage the benefits of a model trained on synthetic data and used for predictions on a different dataset. Finally, to address the challenge of accounting for additional genetic and biomechanical factors, we plan to incorporate models with improved hierarchical relationships (e.g., capsule networks) and quantification of uncertainties in the local tissue mechanical properties \cite{rego2021uncertainty}. Multi-fidelity DeepONet approaches could also be designed, utilizing an additional low-fidelity dataset to drastically reduce the necessary high-fidelity dataset. To make use of the available historical data, we plan to use a Bayesian framework to learn from functional priors (\emph{in silico}) and make posterior estimations (\emph{in vivo}). Such an approach would reduce the use of high-fidelity data required for training, potentially enabling improved prognosis of aneurysm growth and rupture risk in a clinically relevant time frame. 

\subsection{Summary}

\noindent We have developed a novel framework to predict TAA pathology by integrating a constrained mixture model for arterial growth and remodeling with a deep neural network surrogate model. 3D finite element simulations of TAA development arising from randomly distributed losses of elastic fiber integrity and dysfunctional mechanosensing provided synthetic training data for the surrogate model, capable of predicting insult profiles from dilatation and distensibility information. Finally, we demonstrated improved performance using convolutional neural networks in our DeepONet construction. This framework can ultimately be applied to construct patient-specific profiles for aneurysm growth, which will provide critical information to contextualize the predicted mechanobiological insult and forecast the short-term evolution of TAA imaged at one time point in the clinic. Characterization of this progression could play an integral role in determining future patient risk and in designing improved therapeutic interventions.

\newpage
\begin{appendices}

\section{Constrained mixture model}

\subsection{Constitutive properties}
\label{app:appendix_A_constprop}

\noindent The biomechanical behaviors of the structurally significant constituents of the aortic wall $\alpha = e, c, m$ (elastin-dominated matrix, collagen-dominated matrix, and smooth muscle cells, respectively) at current time $s$ are described by strain energy density functions $W_R^\alpha$, with subscript $R$ denoting the reference configuration. The constitutive relations are weighted by their respective volume fractions $\phi^\alpha$. We define $\mathbf{F}$ as the deformation gradient tensor for the tissue from reference to current configurations and $J = \det \mathbf{F}$.

\bigskip
\noindent \textbf{Elastin-dominated matrix.} A neo-Hookean model is used to describe the isotropic elastin-dominated mechanical response ($\alpha = e$), defined as

\begin{equation} \label{eq:elastin}
W_R^e(s) = \phi_R^e(s) \hat{W}^e(s) = \phi_R^e(s) \frac{c^e}{2} (\mathrm{tr} (\mathbf{F}^{e\mathrm{T}}(s) \mathbf{F}(s)) - 3),
\end{equation} \smallskip

\noindent where $\phi_R^e(s) = J\phi^e(s)$ is the referential mass fraction, $\mathbf{F}^e = \mathbf{F}\mathbf{G}^e$, with the tensor $\mathbf{G}^e$ representing the elastin deposition stretch when produced, and $c^e$ is the shear modulus, which can be estimated by fitting biomechanical testing data. Because elastin does not turn over continuously in maturity, its natural configuration is fixed and no convolution integral is required.

\bigskip
\noindent \textbf{Collagen-dominated matrix and passive smooth muscle.} For collagen fibers ($\alpha = c$) and smooth muscle fibers ($\alpha = m$), a four-fiber family model is employed in combination with production and removal, giving the relation

\begin{equation} \label{eq:colsmc}
W_R^{c,m}(s) = \frac{1}{\rho} \int_{-\infty}^s m_R^{c,m}(\tau) q^{c,m}(s,\tau) \hat{W}^{c,m}(\lambda_{n(\tau)}^{c,m}(s)) d\tau,
\end{equation} \smallskip

\noindent where $\rho$ is the mass density of the tissue, $\phi_R^{c,m}(s) = J\phi^{c,m}(s)$, $\lambda_{n(\tau)}^{c,m}(s)$ are the fiber stretches (relative to their evolving natural configurations $n(\tau)$), and

\begin{equation} \label{eq:colsmchat}
\hat{W}^{c,m}(\lambda_{n(\tau)}^{c,m}(s)) = \frac{c_1^{c,m}}{4c_2^{c,m}} \big( \exp \big[ c_2^{c,m} (\lambda_{n(\tau)}^{c,m}(s) - 1)^2 \big] - 1 \big),
\end{equation} \smallskip

\noindent where $c_i^{c,m}$ are material parameters for each fiber family that can be estimated similarly from biomechanical testing data. Collagen fibers are classified into circumferentially, axially, and diagonally oriented (at angle $\alpha_{0}$) populations with associated fractions $\beta^\theta$, $\beta^z$, and $\beta^d$, respectively.

\subsection{Mechanobiologically equilibrated constrained mixture model}
\label{app:appendix_A_mbe}

\noindent In cases of TAA, G\&R can be assumed to reach a quasi-static mechanobiological equilibrium at time $s \gg 0$. Since production balances removal, $m_h^\alpha = m_{Nh}^\alpha = k_{Nh}^\alpha \rho_h^\alpha$, and \autoref{eq:colsmc} reduces to

\begin{equation}
W_{Rh}^{c,m} = \phi_{Rh}^{c,m} \hat{W}^{c,m}(G_h^{c,m}),
\end{equation} \smallskip

\noindent where $G_h^{c,m}$ are the homeostatic deposition stretches of collagen and smooth muscle fibers. Accordingly, rule-of-mixtures expressions can be used to be used for stored energy ($W_{Rh} = \sum \phi_{Rh}^\alpha \hat{W}^\alpha$, where $\phi_{Rh}^\alpha$ are the evolved constituent mass fractions). Similarly, for the Cauchy stresses (\autoref{eq:Cauchystress}),

\begin{equation}
\bm{\sigma}_h = -p_h\mathbf{I} + \sum_{\alpha}^{e,c,m} \phi_h^\alpha \hat{\bm{\sigma}}_h^\alpha,
\end{equation} \smallskip

\noindent where $\hat{\bm{\sigma}}_h^\alpha$ are the constituent-specific Cauchy stresses, and $p_h$ is the equilibrated Lagrange multiplier for the quasi-static G\&R evolution.

We assume the turnover rates of smooth muscle and collagen from the initial to equilibrated states are the equivalent, letting the smooth muscle-to-collagen turnover ratio $\eta = (k^mK_i^m)/(k^cK_i^c) = 1$ ($i = \sigma, \tau_w$), which yields $\rho_{Rh}^m/\rho_o^m = \rho_{Rh}^c/\rho_o^c$. Additionally, assuming that the cardiac output, and thus blood flow rate $Q$, remains constant and laminar, we let $\tau_{wh}/\tau_{wo} = Q a_o^3 / (Q a_h^3)$ and $K_{\tau_w}^\alpha / K_\sigma^\alpha = 0$. We compute the deviation in intramural stress using $\sigma_h = \frac{1}{3} \mathrm{tr} (\bm{\sigma}_h)$ and also allow collagen fibers to gradually reorient during G\&R. Material parameters for the model, shown in \autoref{tab:febio}, are adapted from \cite{Latorre2020b}.

\begin{table}[!htbp]
\centering
\caption{Material parameters used in finite element simulations. Superscripts $e, m, c$ denote elastin-dominated, smooth muscle, and collagen-dominated matrix, respectively; super/subscripts $r, \theta, z, d$ denote radial, circumferential, axial, and symmetric diagonal directions, respectively. Subscript $o$ denotes the initial homeostatic state; subscripts $i = \sigma, \tau_w$ denote intramural and wall shear stress related values, respectively. For elastic fiber integrity insults, the baseline value of $c^e$ is reduced by a prescribed percentage; for mechanosensing insults, $\delta$ is increased. The reader is referred to \cite{Latorre2020b} for further details on these model parameters.}
\resizebox{\textwidth}{!}{
\begin{tabular}{ | l c c | }
\hline
Inner radius, thickness, length & $r_o, h_o, l_o$ & 0.647 mm, 0.040 mm, 15 mm \\
Elastin, smooth muscle, collagen mass fractions & $\phi_o^e, \phi_o^m, \phi_o^c$ & 0.34, 0.33, 0.33 \\
Collagen orientation fractions & $\beta^\theta, \beta^z, \beta^d$ & 0.056, 0.067, 0.877 \\
Diagonal collagen orientation & $\alpha_{0o}$ & $29.9^\circ$ \\
Elastic material parameters & $c^e, c_1^m, c_2^m, c_1^c, c_2^c$ & 89.71 kPa, 261.4 kPa, 0.24, 234.9 kPa, 4.08 \\
Deposition stretches & $G_\theta^e, G_z^e, G_r^e, G^m, G^c$ & 1.90, 1.62, $1/(G_\theta^eG_z^e)$, 1.20, 1.25 \\
Smooth muscle-to-collagen turnover ratio & $\eta$ & 1.0 \\
Shear-to-intramural gain ratio & $K_{\tau_w}/K_\sigma$ & 0.00 \\
Dysfunctional mechanosensing & $\delta$ & 0 \\
\hline
\end{tabular}}
\label{tab:febio}
\end{table}

\section{Random insult fields}
\label{app:appendix_B}

\subsection{Derivation of mean and variance}
\label{app:appendix_B:mean_and_variance}

\noindent Marginally, the values of a GRF with constant mean are themselves normally distributed with the same mean and with variance equal to the overall variance of the GRF. Holding $\varphi$ and $\epsilon$ constant, we may thus solve for the corresponding mean $\mu$ and variance $\varsigma^2$ explicitly. To enforce the overall insult propensity $\varphi$, defined as the fraction of $\vartheta^\ast$ (and thus $\vartheta$) values greater than 0.5, we evaluate the Gaussian complementary CDF at 0.5 and solve for $\mu$ in terms of $\varphi$ and $\varsigma$:

\begin{equation}
    \begin{aligned}
        \varphi &= \frac{1}{2} - \frac{1}{2} \, \mathrm{erf} \left( \frac{0.5 - \mu}{\varsigma \sqrt{2}} \right) \\
        \implies \mu &= \frac{1}{2} - \varsigma \sqrt{2} \, \mathrm{erf}^{-1} \left( 1 - 2 \varphi \right).
    \end{aligned}
    \label{eqn:mean}
\end{equation} \smallskip

\begin{figure}[!htbp]
\centering
\includegraphics[width=0.95\textwidth]{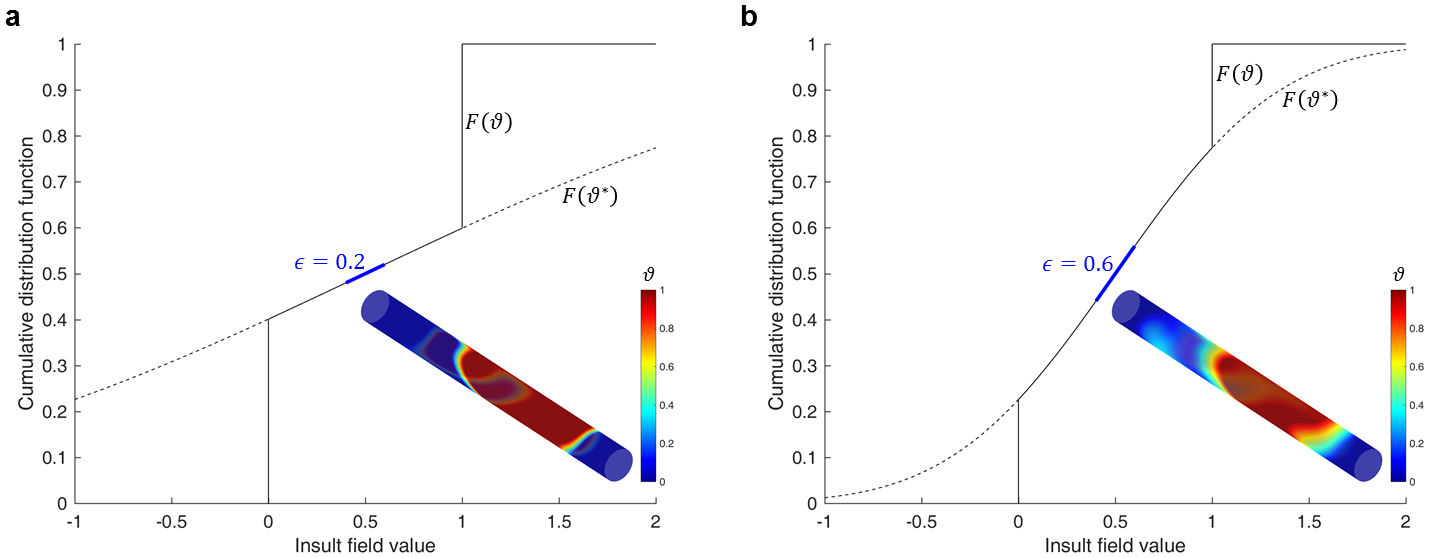}
\caption{CDFs of insult field values before ($\vartheta^\ast$) and after ($\vartheta$) censoring, with insets illustrating the effect of $\epsilon$ on the softness of the boundaries between normal and insult regions ($\varphi = 50$\%, $L_\theta = 1.5$~mm, and $L_z = 2$~mm herein). Smaller values of $\epsilon$ correspond to sharper boundaries, while larger values correspond to softer boundaries.}
\label{fig:insult_cdf}
\end{figure}

\noindent To tune the softness of boundaries between normal and insult regions, we prescribe the slope $\epsilon$ of the CDF of $\vartheta^\ast$ (i.e., its probability density) at $\vartheta^\ast = 0.5$. Note that larger $\epsilon$ values ultimately correspond to more $\vartheta^\ast$ values lying between 0 and 1 (indicating partial insult), and thus softer normal/insult boundaries (\autoref{fig:insult_cdf}). Evaluating the (Gaussian) probability density at 0.5, substituting \autoref{eqn:mean} for $\mu$, and solving for $\varsigma$ yields

\begin{equation}
    \begin{aligned}
        \epsilon &= \frac{1}{\varsigma \sqrt{2 \pi}} \exp \left( -\frac{1}{2} \left( \frac{0.5 - \mu}{\varsigma} \right)^2 \right) \\
        \implies \epsilon &= \frac{1}{\varsigma \sqrt{2 \pi}} \exp \left( -\left[ \mathrm{erf}^{-1} \left( 1 - 2 \varphi \right) \right]^2 \right) \\
        \implies \varsigma &= \frac{1}{\epsilon \sqrt{2 \pi}} \exp \left( -\left[ \mathrm{erf}^{-1} \left( 1 - 2 \varphi \right) \right]^2 \right).
    \end{aligned}
    \label{eqn:standard_deviation}
\end{equation} \smallskip

\noindent Substituting \autoref{eqn:standard_deviation} into \autoref{eqn:mean} and squaring \autoref{eqn:standard_deviation}, we obtain final expressions for the mean and variance,

\begin{equation}
\begin{aligned}
    \mu &= \frac{1}{2} - \frac{1}{\epsilon \sqrt{\pi}} \, \mathrm{erf}^{-1} \left( 1 - 2 \varphi \right) \exp \left( -\left[\mathrm{erf}^{-1} \left( 1 - 2 \varphi \right) \right]^2 \right) \\
    \textrm{and} \quad \varsigma^2 &= \frac{1}{2 \pi \epsilon^2} \exp \left( -2 \left[\mathrm{erf}^{-1} \left( 1 - 2 \varphi \right) \right]^2 \right).
\end{aligned}
\label{eqn:mean_and_variance_final}
\end{equation} \smallskip

\subsection{Sensitivity analysis}
\label{app:appendix_B:sensitivity}

\begin{figure}[!htbp]
\centering
\includegraphics[width=0.95\textwidth]{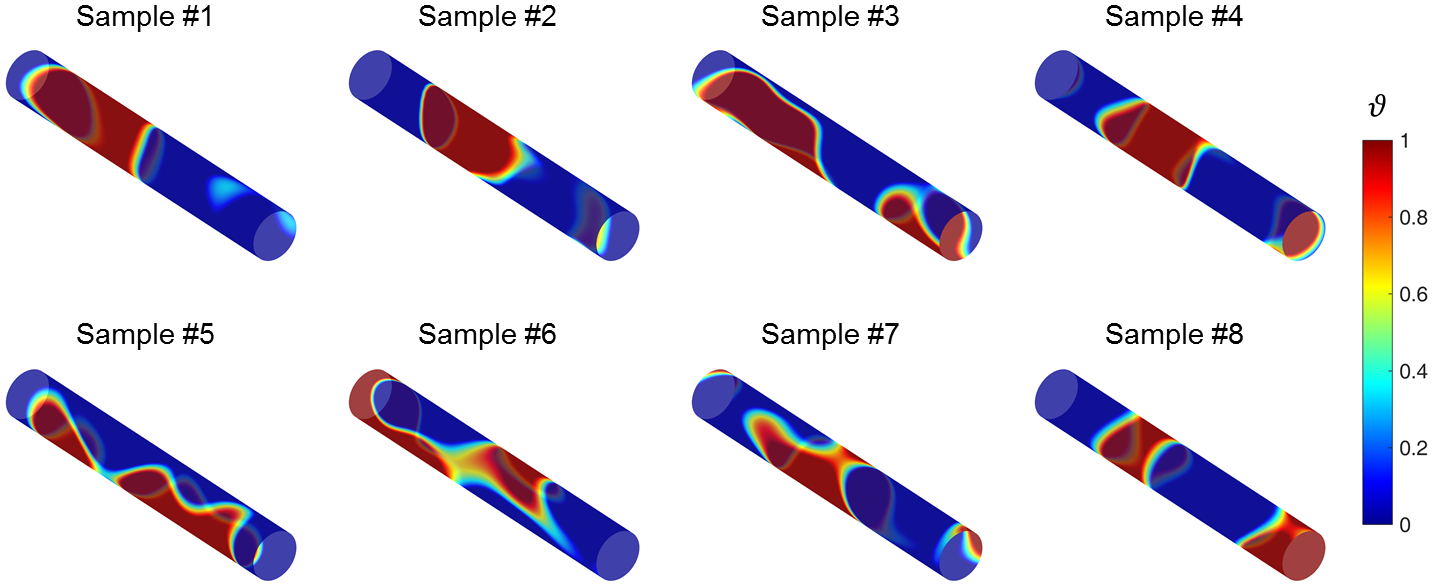}
\caption{Representative random samples generated using the pipeline described in \autoref{sec:methods:modeling_TAA_growth:random_insult_profiles}. Herein, the following were held constant: $\varphi = 35$\%, $\epsilon = 0.2$, $L_\theta = L_z = 2$~mm.}
\label{fig:insult_random_samples}
\end{figure}

\begin{figure}[!htbp]
\centering
\includegraphics[width=0.95\textwidth]{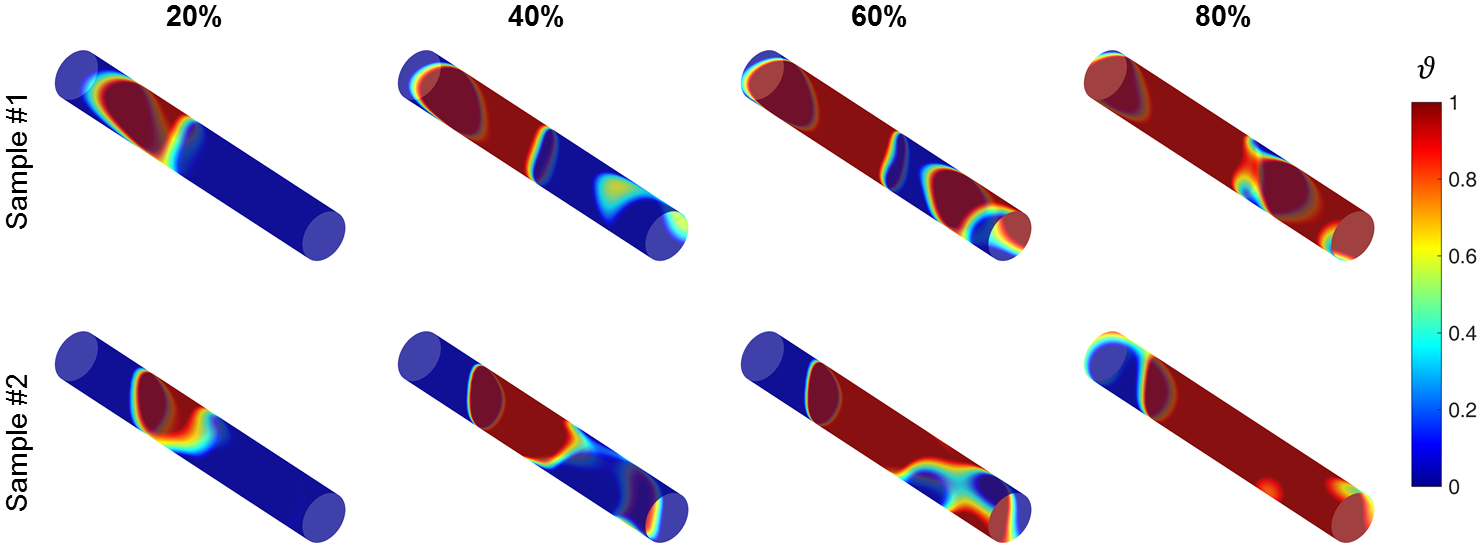}
\caption{Sensitivity of representative random insult profiles with respect to the overall propensity of insult $\varphi$, holding the following constant: $\epsilon = 0.2$, $L_\theta = L_z = 2$~mm.}
\label{fig:insult_propensity}
\end{figure}

\begin{figure}[!htbp]
\centering
\includegraphics[width=0.95\textwidth]{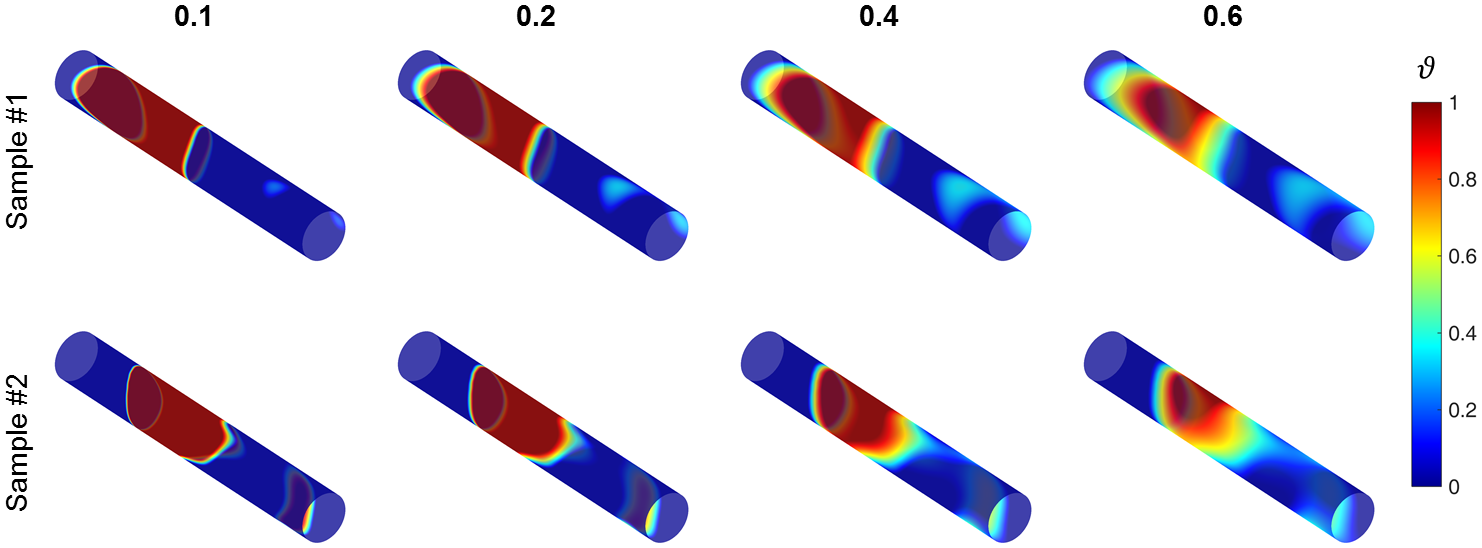}
\caption{Sensitivity of representative random insult profiles with respect to the boundary softness parameter $\epsilon$, holding the following constant: $\varphi = 35$\%, $L_\theta = L_z = 2$~mm.}
\label{fig:insult_boundary_softness}
\end{figure}

\begin{figure}[!htbp]
\centering
\includegraphics[width=0.95\textwidth]{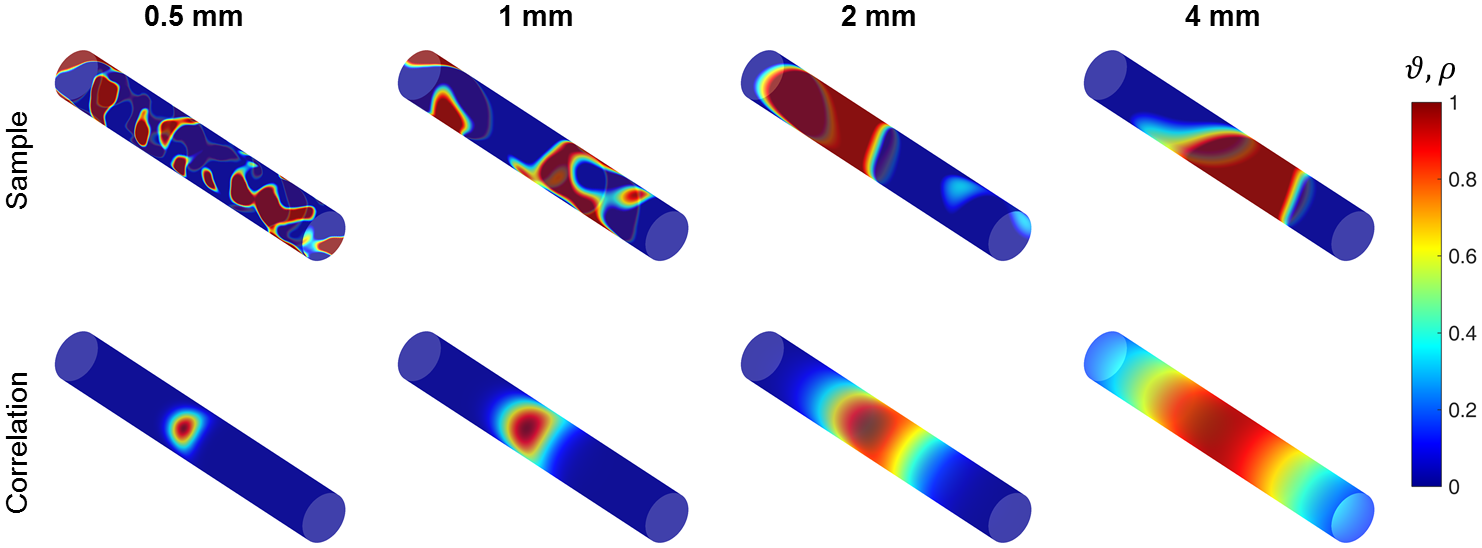}
\caption{(top) Sensitivity of a representative random insult profile with respect to the insult size $L_\theta = L_z$, holding the following constant: $\varphi = 35$\%, $\epsilon = 0.2$. (bottom) Correlation function $\varrho$ evaluated at a central point, showing how the bandwidth of spatial correlation between points increases with increasing insult size. Note that as $L_\theta$ and $L_z$ (and thus also spatial correlation) increase, the insult profile transitions from producing several small insult regions to only a few larger insult regions, while still matching the prescribed overall perturbed area fraction $\varphi$.}
\label{fig:insult_size}
\end{figure}

\begin{figure}[!htbp]
\centering
\includegraphics[width=0.95\textwidth]{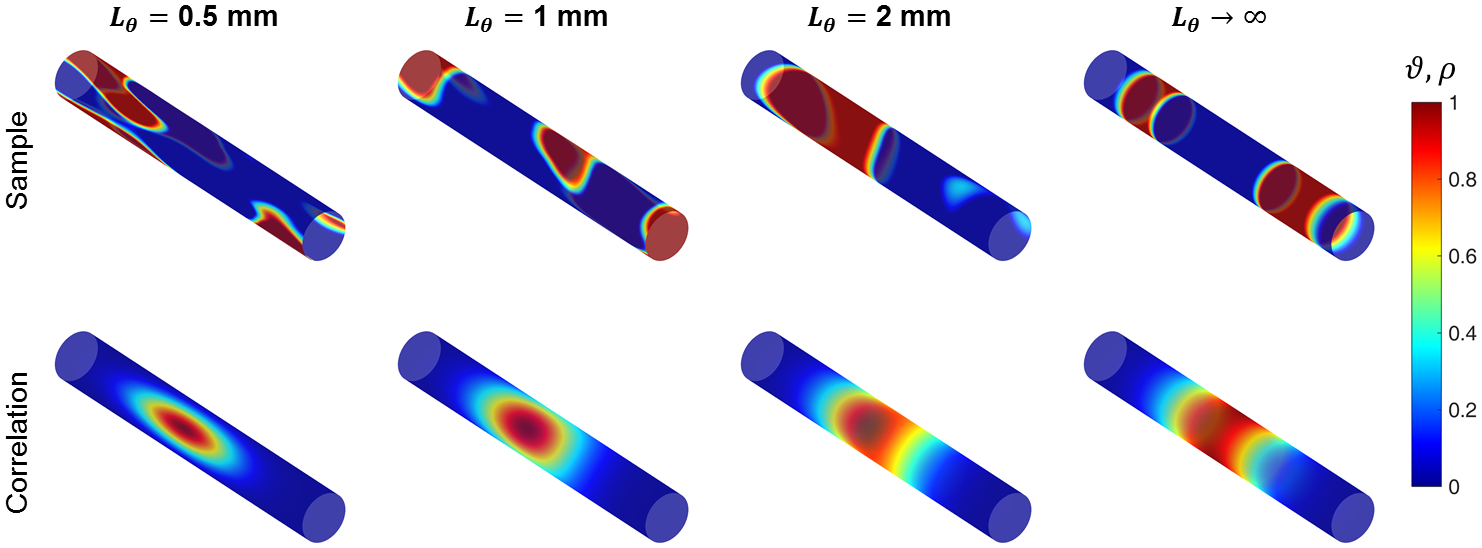}
\caption{(top) Sensitivity of a representative random insult profile with respect to the circumferential insult length scale $L_\theta$, holding the following constant: $\varphi = 35$\%, $\epsilon = 0.2$, $L_z = 2$~mm. (bottom) Correlation function $\varrho$ evaluated at a central point, showing how the bandwidth of spatial correlation between points increases in the circumferential direction with increasing $L_\theta$ while remaining unchanged in the axial direction (since $L_z$ is held constant). In the limit, the generated insult profile is constrained to be perfectly axisymmetric, varying only in the axial direction.}
\label{fig:insult_axial_symmetry}
\end{figure}

\noindent The pipeline detailed in \autoref{sec:methods:modeling_TAA_growth:random_insult_profiles} and \autoref{app:appendix_B:mean_and_variance} is capable of generating random insult profiles that are greatly variable in appearance (\autoref{fig:insult_random_samples}), while precisely controlling desired physically meaningful metrics, including the overall fraction of the vessel experiencing insult ($\varphi$), the softness of the boundaries between normal and insult regions ($\epsilon$), and the size of the insult region(s) in both the circumferential and axial directions ($L_\theta$ and $L_z$, respectively). To demonstrate and quantify how $\varphi$, $\epsilon$, $L_\theta$, and $L_z$ jointly control the size, shape, and appearance of randomly generated insults, we present here a sensitivity analysis illustrating how each of these affect the resulting insult profile. Note that while the insult propensity (\autoref{fig:insult_propensity}) and boundary softness (\autoref{fig:insult_boundary_softness}) parameters alter the insult profile globally while mostly preserving the relative location(s) and shape(s) of the insults, the length scale parameters (Figures~\ref{fig:insult_size} and \ref{fig:insult_axial_symmetry}) affect the profile more locally. When varied in tandem (\autoref{fig:insult_size}), $L_\theta$ and $L_z$ tend to uniformly scale the size of insult regions, conjoining them as needed to preserve the prescribed insult propensity. In contrast, when the length scales are varied individually (\autoref{fig:insult_axial_symmetry}), the insult region size is only altered in the direction of interest, thus tending to change the aspect ratio of the insult regions that are generated. In the limit, as $L_\theta \to \infty$, the insult profile is constrained to be perfectly symmetric about the axial direction, thus guaranteeing that only fusiform aneurysms are produced. The distinction between varying the size and aspect ratio of the insult(s) is highlighted by examining the corresponding GRF correlation function

\begin{equation}
    \varrho(z_o, \theta_o, z_o', \theta_o') = \frac{\kappa(z_o, \theta_o, z_o', \theta_o')}{\varsigma^2} = \exp \left( -\frac{1}{2} \left[ \left( \frac{D_\theta \left( \theta_o, \theta_o' \right)}{L_\theta} \right)^2 + \left( \frac{D_z \left( z_o, z_o' \right)}{L_z} \right)^2 \right] \right),
\end{equation} \smallskip

\noindent which governs the pairwise correlation in $\vartheta^\ast$ values between different points on the vessel wall (Figures~\ref{fig:insult_size} and \ref{fig:insult_axial_symmetry}, bottom panels).

\section{DeepONet model}
\label{app:appendix_c}

\noindent To understand the underlying mathematical foundation of a DNN, we consider a network with $L$ hidden layers, where the $0$-th layer denotes the input layer and the $(L+1)$-th layer is the output layer; the weighted input $\bm z^l_i$ into a $i$\textsuperscript{th} neuron on layer $l$ is a function of weight $\bm W^l_{ij}$ and bias $\bm b^{l-1}_j$ and is represented as

\begin{equation} \label{eq:weighted_input}
    \bm z^l_i =  \mathcal{R}_{l-1}\left(\sum_{j=1}^{m_{l-1}}\left(\bm W^l_{ij}(\bm z^{l-1}_j) + \bm b^{l}_j\right)\right),
\end{equation} \smallskip

\noindent where $\mathcal{R}_{l-1}\left( \cdot \right)$ denotes the activation function of layer $l$, and $m_{l-1}$ is the number of neurons in layer $l-1$.
Based on the foregoing concepts, the feed-forward algorithm for computing the output $\bm{Y}^L$ is expressed as follows:

\begin{equation} \label{eq:feedforward}
\begin{split}
       \bm{Y}^L &= \mathcal{R}_L(\bm W^{L+1}\bm{z}^L + \bm b^L)\\
       \bm{z}^L & = \mathcal{R}_{L-1}\left(\bm{W}^{L}\bm{z}^{L-1} + \bm b^L\right)\\
       \bm{z}^{L-1} & = \mathcal{R}_{L-2}\left(\bm{W}^{L-1}\bm{z}^{L-2} + \bm b^{L-1}\right)\\
        &\;\;\;\;\;\;\;\vdots\\
        \bm z^1 &= \mathcal{R}_0\left(\bm W^1\bm{x} + \bm b^1\right),\\
\end{split}
\end{equation} \smallskip

\noindent where $\bm{x}$ is the input of the neural network. \autoref{eq:feedforward} can be encoded in compressed form as $\bm Y = \mathbb N (x;\bm{\theta})$, where $\bm{\theta} = \left(\bm W, \bm b \right)$ includes both the weights and biases of the neural network $\mathbb N$. Taking into account a DeepONet, the branch network takes as input the function to denote the input realizations $\bm{U} = \{\bm{u}_1, \bm{u}_2, \ldots, \bm{u}_N\}$ for $N$ samples, discretized at $n_{sen}$ sensor locations such that $\bm{u}_i = \{u_i(\bm x_1), u_i(\bm x_2), \ldots, u_i(\bm x_{n_{sen}})\}$ and $i \in [1,N]$. The trunk net inputs the location $\bm y = \{\yb_1,\yb_2,\cdots,\yb_p\}=\{(\hat x_1,\hat y_1),(\hat x_2, \hat y_2), \ldots, (\hat x_p,\hat y_p)\}$ to evaluate the solution operator, where $\hat{x}_i$ and $\hat{y}_i$ denote the coordinates $x$ and $y$ of the point $\yb_i$, respectively. Let us consider that the branch neural network consists of $l_{br}$ hidden layers, where the $(l_{br}+1)$\textsuperscript{th} layer is the output layer consisting of $q$ neurons. Considering an input function $\bm u_{i}$ in the branch network, the network returns a feature embedded in $[b_1, b_2, \ldots, b_q]^\mathrm{T}$ as output. The output $\bm{z}_{br}^{l_{br}+1}$ of the feed-forward branch neural network is expressed as

\begin{equation} \label{eq:output_branch}
    \begin{split}
    \bm{z}_{br}^{l_{br}+1} &= \left[b_1, b_2, \ldots, b_q\right]^\mathrm{T}\\
    &=\mathcal{R}_{br}\left(\bm W^{l_{br}}\bm{z}^{l_{br}} + \bm b^{l_{br}+1}\right),
    \end{split}
\end{equation} \smallskip

\noindent where $\mathcal{R}_{br}\left( \cdot \right)$ denotes the nonlinear activation function for the branch net and $\bm{z}^{l_{br}} = f_{br}(u_i(\bm x_1), u_i(\bm x_2), \ldots, u_i(\bm x_{n_{sen}}))$, where $f_{br}\left(\cdot\right)$ denotes a branch net function. Similarly, consider a trunk network with $l_{tr}$ hidden layers, where the $(l_{tr}+1)$-th layer is the output layer consisting of $q$ neurons. The trunk net outputs a feature embedding $[t_1, t_2, \ldots, t_q]^\mathrm{T}$. The output of the trunk network can be represented as

\begin{equation} \label{eq:output_trunk}
    \begin{split}
    \bm{z}_{tr}^{l_{tr}+1} &= \left[t_1, t_2, \ldots, t_q\right]^\mathrm{T}\\
    &=\mathcal{R}_{tr}\left(\bm W^{l_{tr}}\bm{z}^{l_{tr}} + \bm b^{l_{tr}+1}\right),
    \end{split}
\end{equation} \smallskip

\noindent where $\mathcal{R}_{tr}\left( \cdot \right)$ denotes the non-linear activation function for the trunk net and $\bm{z}^{l_{tr}-1} = f_{tr}(\bm y_1, \bm y_2, \ldots, \bm y_p)$. The key point is that we uncover a new operator $\mathbb G_{\bm{\theta}}$ as a neural network that can infer quantities of interest from unseen and noisy inputs. The two networks are trained to learn the solution operator such that

\begin{equation}
    \mathbb G_{\bm{\theta}}:\bm{u}_i \rightarrow \mathbb G_{\bm{\theta}}(\bm{u}_i),\;\; \forall\;\; i = \{1,2,3, \ldots, N\}.
\end{equation} \smallskip

\noindent For a single input function $\bm u_i$, the DeepONet prediction $\mathbb G_{\bm \theta}(\bm u)$ evaluated at any coordinate $\bm y$ can be expressed as

\begin{equation} \label{eq:output_deeponets}
    \begin{split}
    \mathbb G_{\bm{\theta}}(\bm u_{i})(\bm y) &= \sum_{k = 1}^{q}\left(\mathcal{R}_{br}(\bm W^{l_{br}}_k\bm{z}^{l_{br}-1}_k + \bm b^{l_{br}}_k)\cdot \mathcal{R}_{tr}(\bm W^{l_{tr}}_k\bm{z}^{l_{tr}-1}_k + \bm b^{l_{tr}}_k)\right)\\
    &= \sum_{k = 1}^{q}b_k(u_i(\bm x_1), u_i(\bm x_2), \ldots, u_i(\bm x_m))\cdot t_k(\bm y).
    \end{split}
\end{equation} \smallskip

\noindent DeepONet requires large annotated datasets of paired input-output observations, but it provides a simple and intuitive model architecture that is fast to train, allowing a continuous representation of the target output functions that is resolution-independent. Conventionally, the trainable parameters of the DeepONet represented by $\bm{\theta}$ in \autoref{eq:output_deeponets} are obtained by minimizing a loss function. Common loss functions used in the literature include the $L_1$- and $L_2$-loss functions, defined as

\begin{equation} \label{eq:L1_loss}
\begin{split}
    \mathcal L_1 &= \sum_{i =1}^n \sum_{j =1}^p \big| \mathbb G(\bm u_{i})(\bm y_j) - \mathbb G_{\bm{\theta}}(\bm u_{i})(\bm y_j)\big|\\
    \mathcal L_2 &= \sum_{i =1}^n \sum_{j =1}^p\big(\mathbb G(\bm u_{i}(\bm y_j) - \mathbb G_{\bm{\theta}}(\bm u_{i})(\bm y_j)\big)^2,\\
\end{split}
\end{equation} \smallskip

\noindent where $\mathbb{G}_{\bm{\theta}}(\bm u_{i})(\bm y_j)$ is the predicted value obtained from the DeepONet, and $\mathbb G(\bm u_{i})(\bm y_j)$ is the target value.

\end{appendices}

\newpage
\bibliographystyle{elsarticle-num} 
\bibliography{cas-refs}
\end{document}